\documentclass[a4paper,fleqn]{cas-sc}

\usepackage[authoryear]{natbib}

\usepackage{amssymb}
\usepackage{amsmath}
\usepackage{tikz}
\usepackage{url}
\usepackage{booktabs}
\usepackage{longtable}
\usepackage{chngcntr}
\usepackage{array}
\usepackage{siunitx}
\usepackage{multirow}
\usepackage{makecell}
\usepackage{threeparttable}
\usepackage{graphicx}
\usepackage{subcaption}
\usepackage{tabularx}
\usepackage{float}
\usepackage{placeins}
\usepackage{flafter}

\usetikzlibrary{arrows.meta,positioning,fit,calc}

\begin{document}
	\let\WriteBookmarks\relax
	
	\shorttitle{Gated Concept Adapters}
	
	\shortauthors{Ranjbar et al.}
	
\title[mode=title]{Gated Concept Adapters: Layer-Wise Concept Bottlenecks for Auditable Text Understanding}
	
	
	\author[1]{Niloofar Ranjbar}
	\ead{nranjbar@aut.ac.ir}
	\credit{Conceptualization, Methodology, Formal analysis, Investigation, Writing -- original draft, Writing -- review and editing}
	
	\author[1]{Saeedeh Momtazi}
	\cormark[1]
	\ead{momtazi@aut.ac.ir}
	\credit{Conceptualization, Methodology, Supervision, Validation, Writing -- review and editing}
	
	\author[1]{Mohammad Mehdi Homayounpoor}
	\ead{homayoun@aut.ac.ir}
	\credit{Supervision, Validation, Writing -- review and editing}
	
	\affiliation[1]{
		organization={Department of Computer Engineering, Amirkabir University of Technology (Tehran Polytechnic)},
		city={Tehran},
		country={Iran}
	}
	
	\cortext[1]{Corresponding author}
	
	
\begin{abstract}
	Human-facing text understanding systems are increasingly used to infer affect,
	moderate harmful content, and support socially sensitive decisions. However,
	Transformer predictions are often difficult to audit because they are produced
	from high-dimensional hidden states that do not provide a compact, inspectable
	interface for human analysis or intervention. We introduce \emph{Gated Concept Adapters} (GCA), a layer-wise concept
	bottleneck framework that attaches gated, low-dimensional concept-producing
	readout adapters to selected outputs of pretrained Transformers. Token
	representations are mapped into concept activations, filtered by learned gates,
	pooled into layer-level summaries, and exposed to the prediction head as a
	concept-mediated interface. The resulting dimensions can be inspected through
	usage statistics, signature $n$-grams, post-hoc concept names, and targeted
	concept interventions.
	
	We evaluate GCA on fine-grained emotion classification, toxicity detection,
	topic classification, and a controlled verb-cloze stress test. In the affective
	and safety-relevant classification settings, GCA learns comparatively
	low-redundancy and intervention-sensitive concept representations while
	retaining useful predictive performance. On toxicity detection, it remains close
	to an unconstrained ModernBERT baseline while enabling concept-level inspection,
	targeted intervention, and lightweight post-hoc control. On GoEmotions, the
	end-to-end concept model incurs a macro-F1 cost relative to unconstrained
	RoBERTa, while a task-initialized frozen variant substantially reduces this gap.
	On AG News, GCA remains competitive with recent concept-based baselines, and the
	Qwen verb-cloze experiment clarifies the limits of concept-mediated
	full-vocabulary prediction. Overall, GCA provides a neural concept interface for
	auditable text understanding, prioritizing transparency, diagnosis, and
	intervention over unconstrained accuracy maximization.
\end{abstract}
	
	
	\begin{highlights}
		\item Introduces gated concept adapters for auditable human-centric NLP.
		\item Exposes layer-wise concepts for affect and toxicity prediction.
		\item Tests concept reliance through targeted versus random ablation.
		\item Shows low-redundancy concepts with useful predictive performance.
		\item Supports post-hoc concept inspection and lightweight control.
	\end{highlights}
	
	
	
	\maketitle

\section{Introduction}
\label{sec:introduction}

Pretrained Transformer models are now central components of natural language
processing systems. Encoder models such as RoBERTa and ModernBERT provide strong representations
for text classification \citep{liu2019roberta,modernbert2024}. Decoder-only
models such as Qwen support full-vocabulary token prediction and
language-modeling applications
\citep{qwen2techreport2024,qwen25techreport2024}.
These models are increasingly used in human-facing text understanding settings,
including affect recognition, harmful-content moderation, and socially
sensitive decision-support workflows. In such settings, predictive performance
alone is often insufficient: users, developers, and auditors may also need to
inspect which textual evidence supports a prediction and test how the output
changes under controlled intervention \citep{arrieta2020explainable,
	borkan2019nuanced,song2023abusivebias}.

Despite their predictive strength, Transformer decisions remain difficult to
audit. Predictions are typically produced from high-dimensional hidden states
whose individual coordinates do not correspond to stable, human-understandable
evidence. This opacity is especially problematic for affective and
safety-relevant text understanding, where a model may appear accurate in
aggregate while relying on brittle, redundant, or socially undesirable cues. A
useful neural interface for such systems should therefore expose intermediate
variables that are compact enough to inspect, structured enough to diagnose, and
directly connected to the prediction path so that interventions on those
variables have measurable effects.

Concept-based interpretability addresses this need by explaining model behavior
through intermediate variables that summarize meaningful evidence. Post-hoc
methods such as TCAV, Network Dissection, and ACE identify concept directions,
neuron--concept alignments, or activation clusters in trained networks
\citep{kim2018tcav,bau2017network,ghorbani2019ace}. These methods are valuable
for diagnosis, but the discovered concepts are not necessarily used by the
prediction head. A concept can therefore be descriptively meaningful without
being a variable that functionally mediates the model's output. Concept
Bottleneck Models (CBMs) address this limitation by placing concept variables
between representation and prediction, enabling direct concept-level inspection
and intervention \citep{koh2020concept}. However, classical CBMs typically
require a predefined concept inventory and concept-level supervision, which are
difficult to obtain for open-ended text domains where relevant evidence can be
task-specific, compositional, and context-dependent.

Recent NLP and language-model work has reduced this dependence on manual concept
engineering through textual bottlenecks, concept-bottleneck LLMs, label-free
concept generation, automatically named concept layers, and sparse feature
discovery \citep{Ludan2023TBM,bhan-etal-2025-towards,ijcai2025p1221,
	sun2025concept,oikarinen2023label,Yang2023LaBo,bidusa2025conceptlayers,
	bricken2023monosemantic,cunningham2024sae,huben2024sae}. These directions are
promising, but several limitations remain. Concept sets may be incomplete or
mismatched to the task; post-hoc concepts may not lie on the prediction path;
single-stage or final-layer bottlenecks provide limited visibility into how
evidence is distributed across Transformer depth; and generic adaptation
methods such as adapters, LoRA, and prefix tuning improve task adaptation
without exposing auditable concept variables
\citep{pmlr-v97-houlsby19a,hu2022lora,
	li2021prefixtuningoptimizingcontinuousprompts}. These limitations motivate a
concept interface that is learned inside the Transformer, exposed across layers,
used by the prediction head, and evaluated not only by accuracy but also by
concept structure, post-hoc name grounding, and targeted intervention behavior.

We introduce \emph{Gated Concept Adapters} (GCA), a layer-wise concept
bottleneck framework for pretrained Transformers. GCA attaches concept-producing readout adapters to selected Transformer layer
outputs. Token
representations are projected into low-dimensional concept activations,
filtered by learned gates, pooled into layer-level concept summaries, and passed
to the prediction head as a concept-mediated interface. Thus, GCA is a
\emph{prediction-path bottleneck}: the final prediction is mediated by explicit
concept variables that can be inspected and intervened on. It is not a strict
sequential bottleneck that forces every intermediate hidden state through
concepts before the next Transformer layer. This design preserves the
expressive pretrained backbone while making the prediction-facing path more
auditable.

GCA does not use human concept annotations during training. Instead, it combines
the task loss with modular concept-structure regularizers that encourage
selective activation, reduce redundancy, and improve token-level coherence, with
redundancy reduction motivated by prior work on decorrelation and representation
structure \citep{cogswell2016reducingoverfittingdeepnetworks,kornblith2019cka,
	zbontar2021barlow,bardes2022vicreg}. After training, concept dimensions are
named post hoc by mining signature $n$-grams and representative examples and by
using an LLM-assisted proposal and referee pipeline. These names are treated as
evidence-grounded descriptors for human inspection, not as ground-truth concept
supervision.

A central part of our evaluation is intervention. For each prediction, we
identify concept dimensions that most support the target output and ablate them
at test time. We compare this targeted ablation with random ablation of the same
size. This protocol does not prove that every dimension is a clean semantic
concept or establish complete causal mediation. Rather, it tests whether the
prediction head functionally relies on the selected concept dimensions. This
distinction is important: interpretability claims should not rest only on
plausible names or visualizations, but also on whether the exposed variables are
actually used by the model.

We evaluate GCA in four settings. The primary human-facing evaluations are
GoEmotions for fine-grained multi-label affective text classification
\citep{demszky2020goemotions} and Jigsaw/Civil Comments toxicity detection with
ModernBERT \citep{borkan2019nuanced,modernbert2024}. These tasks test whether
GCA can expose auditable concept pathways in affective and safety-relevant text
understanding. We also evaluate AG News for topic classification and comparison
with CB-LLM and HI-Concept
\citep{zhang2015character,sun2025concept,zhao2024hiconcept}, and a controlled
Qwen verb-cloze stress test based on Universal Dependencies English EWT
\citep{nivre2020universal,silveira2014gold}. AG News provides a compact
single-label benchmark with external concept-based baselines, while the Qwen
experiment tests whether a decoder-side concept-mediated path can remain useful
in a bounded full-vocabulary setting. The Qwen experiment is not intended to
solve open-ended generation.

The results show that GCA provides an accuracy--auditability trade-off rather
than a universal replacement for unconstrained fine-tuning. On GoEmotions, the
end-to-end concept model incurs a macro-F1 cost relative to RoBERTa, especially
on rare labels, but exposes lower-redundancy and intervention-sensitive concept
pathways. A task-initialized frozen variant substantially reduces this gap. On
toxicity detection, GCA remains close to a ModernBERT baseline on the main
toxicity target while enabling concept inspection, targeted intervention, and
lightweight post-hoc control. On AG News, GCA is competitive with recent
concept-based baselines while preserving internally learned layer-wise concept
routing. On Qwen verb cloze, the matched no-concept baseline retains stronger
full-vocabulary ranking performance, but the concept-mediated variant shows
strong targeted-versus-random intervention effects. Overall, GCA is most useful
when inspection, diagnosis, and intervention are important enough to justify
some loss of unconstrained predictive flexibility.

The main contributions are:
\begin{itemize}
	\item We introduce layer-wise gated concept adapters that place concept
	summaries on the Transformer prediction path.
	\item We train the concept interface with modular regularizers for
	selectivity, redundancy reduction, token-level coherence, and
	task-dependent usage control.
	\item We evaluate GCA on affective, safety-relevant, topic-classification,
	and controlled cloze settings against unconstrained fine-tuning,
	adapter-only controls, single-bottleneck variants, task-initialized frozen
	variants, and external concept baselines where available.
	\item We analyze the learned concepts through concentration and redundancy
	diagnostics, post-hoc naming audits, layer-wise routing, and
	targeted-versus-random interventions.
	\item We include a controlled decoder-side Qwen verb-cloze stress test to
	clarify the scope and limits of concept-mediated full-vocabulary
	prediction.
\end{itemize}

The remainder of the paper is organized as follows.
Section~\ref{sec:related_work} reviews related work.
Section~\ref{sec:method} describes GCA.
Section~\ref{sec:experimental_setup} presents datasets, baselines, and metrics.
Section~\ref{sec:results} reports the results.
Section~\ref{sec:discussion} discusses scope and limitations, and
Section~\ref{sec:conclusion} concludes.
\section{Related Work}
\label{sec:related_work}

Gated Concept Adapters (GCA) build on work in concept-based interpretability,
concept bottleneck models, language-model concept discovery, automatic concept
naming, sparse and structured representation learning, adapter-based
fine-tuning, and intervention-oriented auditing. We review these areas with an
emphasis on human-facing text understanding, where affective and safety-relevant
predictions require not only high predictive performance, but also inspectable
evidence, diagnosis of model behavior, and controllable intervention mechanisms.

\subsection{Concept-based interpretability and concept bottlenecks}

Concept-based interpretability explains model behavior through intermediate
factors that are more meaningful than individual hidden units. Post-hoc methods
such as TCAV, Network Dissection, and ACE analyze trained networks by
identifying concept directions, neuron--concept alignments, or coherent
activation clusters
\citep{kim2018tcav,bau2017network,ghorbani2019ace}. These methods are useful
for diagnosis because they connect internal representations to human-readable
patterns. However, the discovered concepts are not necessarily part of the
model's decision pathway. A concept can therefore be descriptively meaningful
without being a variable that the prediction head actually uses.

Concept Bottleneck Models (CBMs) address this limitation by placing concept
variables between the input representation and the final prediction
\citep{koh2020concept}. This design enables direct concept-level inspection and
intervention: changing a concept value can change the downstream prediction.
This property is especially relevant for human-facing systems, where users or
auditors may need to test whether a prediction depends on appropriate evidence.
However, classical CBMs typically require a predefined concept inventory and
concept-level supervision. These assumptions are difficult to satisfy in text
domains, where useful concepts are often task-specific, compositional,
context-dependent, and hard to enumerate in advance. Prior work has also shown
that bottleneck design involves trade-offs among predictive performance,
concept completeness, and intervention reliability \citep{oikarinen2023label}.

GCA follows the intervention-oriented motivation of CBMs but avoids manual
concept supervision. The concept dimensions are learned from task supervision
and regularization, then described post hoc using mined textual evidence. GCA
also exposes concept summaries across multiple Transformer layers rather than
relying only on a final-layer bottleneck. This makes it possible to inspect both
which concepts support a prediction and where in the Transformer depth those
concepts are used.

\subsection{Concept bottlenecks and concept discovery in language models}

Recent work has extended concept bottlenecks and concept discovery to NLP and
large language models. Text Bottleneck Models use natural-language concepts as
intermediate variables for text classification, often with concepts proposed or
refined by language models \citep{Ludan2023TBM}. Other textual CBM variants
seek to reduce manual concept engineering by discovering, selecting, or refining
concept sets automatically
\citep{bhan-etal-2025-towards,ijcai2025p1221}. Concept Bottleneck LLMs study
how concept variables can support classification or control in larger language
models \citep{sun2025concept}, while concept-layer approaches insert
interpretable intermediate representations into language-model pipelines,
sometimes using external knowledge or ontology-guided concept selection
\citep{bidusa2025conceptlayers}.

These approaches are closely related to our goal of making Transformer
predictions more auditable through intermediate concept variables. The main
difference is that many existing methods rely on explicit concept inventories,
LLM-proposed concept sets, or a single concept stage near the output. Such
designs can be effective when the concept set is well matched to the task, but
they may provide limited visibility into how evidence is distributed across
Transformer depth. GCA instead learns latent concept dimensions internally,
exposes them at selected Transformer layers, and places their pooled summaries
on the prediction-facing interface.

We compare with external concept-based baselines where a direct task overlap is
available. AG News provides a compact topic-classification setting for
comparison with CB-LLM \citep{sun2025concept} and HI-Concept
\citep{zhao2024hiconcept}. We interpret this comparison as a trade-off analysis
among different concept-construction regimes, not as evidence that one method
universally dominates the others.

\subsection{Automatic concept naming and evidence grounding}

A related challenge is assigning human-readable names to learned concept
dimensions. Automatic naming methods such as CLIP-Dissect and language-guided
bottlenecks use external models to generate textual descriptions of internal
units or candidate concepts
\citep{oikarinen2023label,Yang2023LaBo}. Such methods make concept analysis
more scalable, but they also introduce a risk: generated names can sound
plausible even when the supporting evidence is weak. This risk is important in
auditable systems because a convincing but poorly grounded concept name can
mislead users about what the model has learned.

For this reason, GCA separates concept learning from concept naming. The model
is not trained with LLM-generated concept labels. After training, we mine
signature $n$-grams and representative examples, use an LLM-assisted proposal
and referee pipeline to assign compact names, and audit whether those names are
grounded in the displayed evidence. Thus, post-hoc names are treated as
diagnostic summaries rather than ground-truth semantic annotations. The naming
audit evaluates whether the evidence supports the assigned descriptor, while
intervention experiments evaluate whether the underlying concept dimension is
functionally used by the prediction head.

\subsection{Sparse representations, redundancy reduction, and adapters}

Work in mechanistic interpretability has shown that neural activations can often
be decomposed into sparse and more inspectable features. Sparse autoencoders
and dictionary-learning methods learn feature dictionaries over Transformer
activations, and some learned features correspond to coherent patterns that are
easier to inspect than raw hidden coordinates
\citep{bricken2023monosemantic,cunningham2024sae,huben2024sae}. These methods
are typically post hoc: they analyze representations of a trained or frozen
model rather than constraining the supervised prediction pathway itself.

GCA is complementary to this direction. It does not learn a post-hoc dictionary
over frozen activations. Instead, it attaches concept-producing readout modules to selected layer outputs and trains the prediction head to depend on the resulting concept
summaries. This makes the learned dimensions available for prediction-time
intervention. The sparse-feature literature nevertheless motivates our emphasis
on selectivity and redundancy reduction: an auditable concept interface should
not spread the same evidence across many correlated dimensions. This motivation
is also related to decorrelation and representation-structure objectives such
as DeCov, CKA-based analysis, Barlow Twins, and VICReg
\citep{cogswell2016reducingoverfittingdeepnetworks,kornblith2019cka,
	zbontar2021barlow,bardes2022vicreg}. In GCA, redundancy reduction is used as a
practical design principle and diagnostic criterion, not as a claim that concept
dimensions are statistically independent.

GCA is also related to adapter-based and parameter-efficient fine-tuning.
Classic adapters insert small trainable modules into Transformer layers
\citep{pmlr-v97-houlsby19a}, while LoRA and prefix tuning adapt pretrained
models through low-rank updates or continuous prompts
\citep{hu2022lora,li2021prefixtuningoptimizingcontinuousprompts}. These methods
are effective for task adaptation, but their trainable channels are not designed
to serve as human-inspectable concept variables. GCA can be viewed as an
adapter-style architecture with a different objective: its modules produce
explicit concept activations, and the prediction head operates on pooled
concept summaries. The evaluation therefore includes not only predictive
performance, but also sparsity, redundancy, layer usage, post-hoc naming, and
intervention sensitivity.

\subsection{Intervention-oriented auditing for human-facing text systems}

Interpretability is most useful when it supports concrete tests of model
behavior. Affective and socially sensitive text understanding provides a natural
setting for such intervention-oriented evaluation. In emotion classification,
labels are fine-grained, overlapping, and often expressed through short
contextual phrases rather than isolated keywords \citep{demszky2020goemotions}.
In toxicity detection, models may rely on identity mentions, pragmatic context,
or strongly negative language in ways that can produce subgroup errors or false
positives. These settings therefore require audit tools that expose not only
which prediction is made, but also which intermediate evidence supports it and
whether that evidence can be modified without retraining the full model.

The Jigsaw/Civil Comments benchmark was designed to expose unintended bias and
subgroup behavior in toxicity classification \citep{borkan2019nuanced}. More
recent work likewise emphasizes the need to evaluate robustness and bias in
abusive-language and moderation systems \citep{song2023abusivebias}. These
concerns motivate evaluation protocols that go beyond aggregate accuracy and
test whether a model's internal evidence can be inspected, perturbed, or
controlled.

Our evaluation follows this audit-oriented view. We use targeted-versus-random
concept ablation to test whether selected concept dimensions are functionally
involved in prediction. This is not treated as proof that every dimension is a
clean semantic concept, nor as evidence of complete causal mediation. Instead,
it is a practical reliance test: if removing the most supportive dimensions
changes the output much more than removing random dimensions of the same size,
then the prediction head is using the concept interface in a nontrivial way. In
the toxicity experiment, we further test whether the learned concept space can
serve as a compact post-hoc control surface. This experiment is not a complete
fairness solution; rather, it evaluates whether concept-level gates can modify
some identity-related sensitivity without retraining the encoder.

\subsection{Positioning of GCA}

GCA combines ideas from concept bottlenecks, adapter-based tuning, sparse
representation learning, automatic concept naming, and intervention-oriented
auditing. Compared with post-hoc concept methods such as TCAV or ACE, GCA places
the concept representation on the prediction path. Compared with classical CBMs,
it does not require predefined concept labels during training. Compared with
many textual bottleneck and CB-LLM-style methods, it emphasizes internally
learned, layer-wise concept routing inside pretrained Transformers rather than a
single concept stage near the output. Compared with generic adapters, LoRA, and
prefix tuning, its trainable modules are designed to expose auditable concept
dimensions rather than only improve adaptation efficiency.

This positioning also clarifies the main limitation. Because GCA learns latent
dimensions from task supervision rather than manually specified concepts, not
every dimension is guaranteed to have a clean human name. Post-hoc labels should
therefore be read as evidence-grounded summaries, not as definitive semantic
annotations. The contribution of GCA is to provide a structured,
low-redundancy, and intervention-sensitive concept interface for
Transformer-based text understanding, especially in settings where affective or
safety-relevant predictions must be inspected, diagnosed, and controlled.

\begin{figure}[pos=t]
	\centering
	\includegraphics[width=0.95\linewidth]{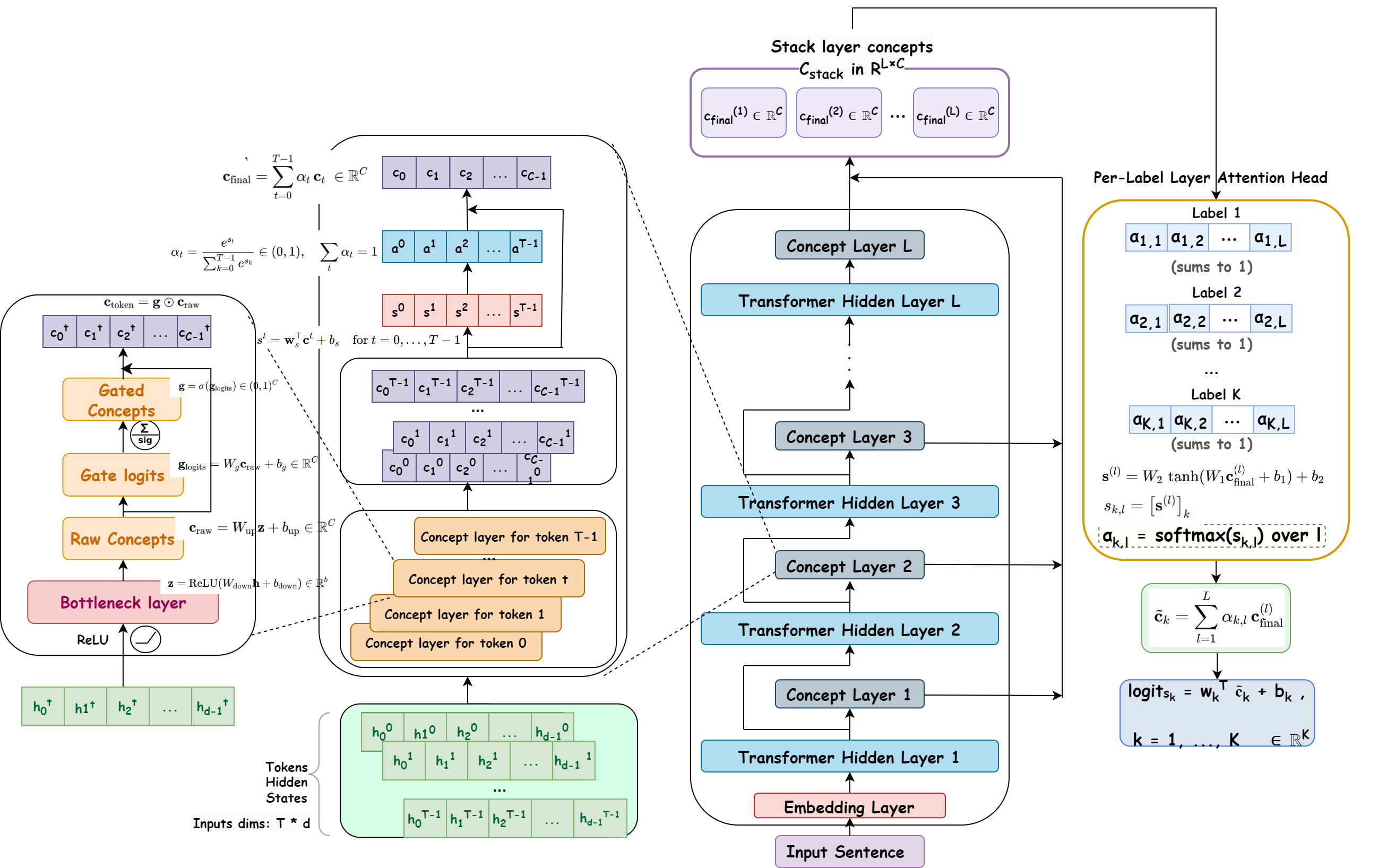}
	\caption{Overview of the proposed Gated Concept Adapter (GCA) framework.
		Concept-producing readout adapters are attached to selected Transformer layer
		outputs. They produce gated token-level concept activations, which are pooled
		into layer-level concept summaries. For classification, a label-wise
		layer-attention head combines these summaries and predicts from the fused
		concept representation. For Qwen verb cloze, the concept stack is projected
		to a decoder-hidden representation and scored with the vocabulary projection.
		The Transformer hidden stream is not replaced by the concept variables;
		instead, the task head predicts from the concept-mediated path.}
	\label{fig:concept-arch}
\end{figure}
\section{Methodology}
\label{sec:method}

\subsection{Problem setting and overview}

We study supervised text understanding and controlled cloze prediction with a
pretrained Transformer backbone augmented by \emph{Gated Concept Adapters}
(GCA). The goal is to retain useful predictive performance while exposing a
concept-mediated prediction path that can be inspected, named post hoc, and
intervened on. Figure~\ref{fig:concept-arch} summarizes the framework.

For classification, an input sequence
$x=(x_1,\ldots,x_T)$ with attention mask $m\in\{0,1\}^{T}$ is mapped to either a
single class label $y\in\{1,\ldots,K\}$ or a multi-label target
$y\in\{0,1\}^{K}$. For the controlled verb-cloze setting, one
\textsc{verb} or \textsc{aux} token is replaced by \texttt{[BLANK]}, and the
model predicts the missing token over the full vocabulary. In both cases, the
prediction head receives concept summaries rather than the original Transformer
hidden states directly. Thus, GCA is a \emph{prediction-path bottleneck}: it
does not force every intermediate hidden state through concepts, but it makes
the final prediction interface concept-mediated and directly testable through
interventions.

The GCA modules act as concept-producing readout adapters attached to selected
Transformer layer outputs. They do not replace or overwrite the hidden states
passed to subsequent Transformer layers. Instead, they create a separate
concept-mediated prediction-facing path from selected hidden states to the task
head. This distinction is important: GCA constrains the information available to
the final prediction head, while preserving the expressive capacity of the
pretrained backbone.

Let $B$ be the batch size, $T$ the sequence length, $d$ the Transformer hidden
size, and $C$ the number of concept dimensions per selected layer. Let
$\mathcal{S}=\{s_1,\ldots,s_L\}$ denote the selected Transformer layers. We
re-index these selected layers by $\ell=1,\ldots,L$, so that
$H^{(\ell)}$ denotes the hidden states extracted from backbone layer $s_\ell$.
For an input batch, the backbone produces hidden states
\begin{equation}
	H^{(\ell)}
	=
	[h_1^{(\ell)},\ldots,h_T^{(\ell)}]
	\in \mathbb{R}^{B\times T\times d},
	\qquad
	\ell=1,\ldots,L .
\end{equation}
For readability, the following equations omit the batch index.

\subsection{Layer-wise gated concept adapters}

At each selected layer $\ell$, GCA maps each token hidden state
$h_t^{(\ell)}\in\mathbb{R}^{d}$ into a $C$-dimensional concept vector. In the
encoder-based classification experiments, we use a bottleneck readout adapter:
\begin{equation}
	z_t^{(\ell)}
	=
	\mathrm{ReLU}
	\left(
	W_{\downarrow}^{(\ell)}h_t^{(\ell)}
	+
	b_{\downarrow}^{(\ell)}
	\right),
	\qquad
	z_t^{(\ell)}\in\mathbb{R}^{r},
\end{equation}
where $r\ll d$,
$W_{\downarrow}^{(\ell)}\in\mathbb{R}^{r\times d}$, and
$b_{\downarrow}^{(\ell)}\in\mathbb{R}^{r}$. The raw concept vector is
\begin{equation}
	c_{\mathrm{raw},t}^{(\ell)}
	=
	W_{\uparrow}^{(\ell)}z_t^{(\ell)}
	+
	b_{\uparrow}^{(\ell)},
	\qquad
	c_{\mathrm{raw},t}^{(\ell)}\in\mathbb{R}^{C},
	\label{eq:raw_concept}
\end{equation}
where $W_{\uparrow}^{(\ell)}\in\mathbb{R}^{C\times r}$ and
$b_{\uparrow}^{(\ell)}\in\mathbb{R}^{C}$. The bottleneck limits adapter capacity
and discourages the concept module from simply copying the full hidden state.

To encourage token-specific concept selection, we compute a gate for each
concept dimension:
\begin{equation}
	g_t^{(\ell)}
	=
	\sigma
	\left(
	\frac{
		W_g^{(\ell)}c_{\mathrm{raw},t}^{(\ell)}
		+
		b_g^{(\ell)}
	}{\tau}
	\right),
	\qquad
	g_t^{(\ell)}\in(0,1)^C ,
	\label{eq:concept_gate}
\end{equation}
where $W_g^{(\ell)}\in\mathbb{R}^{C\times C}$,
$b_g^{(\ell)}\in\mathbb{R}^{C}$, and $\tau>0$ is the gate temperature. The
gated token-level concept vector is
\begin{equation}
	c_t^{(\ell)}
	=
	g_t^{(\ell)}
	\odot
	c_{\mathrm{raw},t}^{(\ell)} .
	\label{eq:gated_concept}
\end{equation}
The gate allows a concept channel to be suppressed for one token while remaining
available for other inputs. In the reported experiments, we use soft gates for
stable end-to-end optimization.

For the decoder-side Qwen verb-cloze experiment, we use a lighter local readout
for computational stability in the full-vocabulary setting. Raw concepts and
gates are computed directly from decoder hidden states:
\begin{equation}
	c_{\mathrm{raw},t}^{(\ell)}
	=
	W_{c}^{(\ell)}h_t^{(\ell)}
	+
	b_{c}^{(\ell)},
	\qquad
	g_t^{(\ell)}
	=
	\sigma
	\left(
	\frac{
		W_{g,h}^{(\ell)}h_t^{(\ell)}
		+
		b_{g,h}^{(\ell)}
	}{\tau}
	\right).
	\label{eq:qwen_direct_concepts}
\end{equation}
The gated concept vector is then computed as in
Eq.~\eqref{eq:gated_concept}. This variant preserves the same
concept-mediated prediction-path constraint, but it is not intended as a full
open-ended generation architecture.

\subsection{Pooling and concept-based prediction}

Each selected layer produces token-level concepts
$\{c_t^{(\ell)}\}_{t=1}^{T}$. We summarize them with mask-aware attention pooling
over valid token positions $\mathcal{V}=\{t:m_t=1\}$. The pooling scores are
computed from the gated token concepts:
\begin{equation}
	s_t^{(\ell)}
	=
	w_s^{(\ell)\top}c_t^{(\ell)}
	+
	b_s^{(\ell)},
	\qquad
	\alpha_t^{(\ell)}
	=
	\frac{\exp(s_t^{(\ell)})}
	{\sum_{j\in\mathcal{V}}\exp(s_j^{(\ell)})},
\end{equation}
where padding positions are excluded by the attention mask. The layer-level
concept summary is
\begin{equation}
	c_{\mathrm{final}}^{(\ell)}
	=
	\sum_{t\in\mathcal{V}}
	\alpha_t^{(\ell)}c_t^{(\ell)},
	\qquad
	c_{\mathrm{final}}^{(\ell)}\in\mathbb{R}^{C}.
	\label{eq:layer_concept_pooling}
\end{equation}
The layer summaries are stacked as
\begin{equation}
	C_{\mathrm{stack}}
	=
	\left[
	c_{\mathrm{final}}^{(1)},
	c_{\mathrm{final}}^{(2)},
	\ldots,
	c_{\mathrm{final}}^{(L)}
	\right]
	\in\mathbb{R}^{L\times C}.
	\label{eq:concept_stack}
\end{equation}
This layer-wise stack exposes how evidence is distributed across Transformer
depth, rather than forcing all concept evidence to appear only at the final
layer.

For classification, we use a label-conditioned layer-attention head. For each
layer $\ell$, a small scoring network produces label-specific scores:
\begin{equation}
	q^{(\ell)}
	=
	W_2
	\tanh
	\left(
	W_1c_{\mathrm{final}}^{(\ell)}
	+
	b_1
	\right)
	+
	b_2,
	\qquad
	q^{(\ell)}\in\mathbb{R}^{K}.
\end{equation}
Dropout is applied inside the layer-scoring network during training and is
omitted from the equations for readability. The attention weight for label $k$
at layer $\ell$ is
\begin{equation}
	a_{k,\ell}
	=
	\frac{\exp(q_k^{(\ell)})}
	{\sum_{j=1}^{L}\exp(q_k^{(j)})}.
	\label{eq:label_layer_attention}
\end{equation}
The fused concept vector for label $k$ is
\begin{equation}
	\tilde{c}_k
	=
	\sum_{\ell=1}^{L}
	a_{k,\ell}
	c_{\mathrm{final}}^{(\ell)},
	\qquad
	\tilde{c}_k\in\mathbb{R}^{C}.
	\label{eq:fused_label_concept}
\end{equation}
The corresponding logit is
\begin{equation}
	o_k
	=
	w_k^\top\tilde{c}_k
	+
	b_k ,
	\label{eq:concept_logit}
\end{equation}
where $w_k\in\mathbb{R}^{C}$ and $b_k\in\mathbb{R}$ are the classifier
parameters for label $k$. Single-layer controls are recovered as the special
case $L=1$.

For Qwen verb cloze, the model must score the full vocabulary rather than a
fixed label set. The input prompt contains the sentence with a \texttt{[BLANK]}
marker and does not include the gold target token. The model predicts the
missing single-token \textsc{verb} or \textsc{aux} target over the full Qwen
vocabulary. We flatten the layer-wise concept stack and project it back to the
decoder hidden size:
\begin{equation}
	h_{\mathrm{concept}}
	=
	\phi
	\left(
	W_{\mathrm{gen}}
	\mathrm{vec}(C_{\mathrm{stack}})
	+
	b_{\mathrm{gen}}
	\right),
	\label{eq:qwen_concept_hidden}
\end{equation}
where $\phi$ is a nonlinear projection and
$\mathrm{vec}(C_{\mathrm{stack}})\in\mathbb{R}^{LC}$. Vocabulary logits are then
computed as
\begin{equation}
	o_v
	=
	e_v^\top h_{\mathrm{concept}}
	+
	b_v,
	\qquad
	v\in\mathcal{V}_{\mathrm{vocab}}.
	\label{eq:qwen_vocab_logits}
\end{equation}
The vectors $e_v$ correspond to the vocabulary projection used for Qwen token
scoring, so the output space remains the original decoder vocabulary. Thus, the
vocabulary distribution is produced from a concept-derived hidden
representation rather than directly from the decoder hidden state.

\subsection{Training objective}

For single-label classification and Qwen verb cloze, we use cross-entropy. For
multi-label classification, we use binary cross-entropy with logits. Class
weighting, threshold tuning, and task-specific choices are reported in
Section~\ref{sec:experimental_setup}.

The task loss alone does not guarantee a selective or low-redundancy concept
space. We therefore train GCA with a modular concept-structured objective:
\begin{equation}
	\label{eq:total_loss}
	\begin{aligned}
		\mathcal{L}
		&=
		\mathcal{L}_{\mathrm{task}}
		+
		\lambda_{1}\mathcal{L}_{1}
		+
		\lambda_{\mathrm{excl}}\mathcal{L}_{\mathrm{excl}}
		+
		\lambda_{\mathrm{gate}}\mathcal{L}_{\mathrm{gate}}
		+
		\lambda_{\mathrm{orth}}\mathcal{L}_{\mathrm{orth}}
		+
		\lambda_{\mathrm{intra}}\mathcal{L}_{\mathrm{intra}}
		+
		\lambda_{\mathrm{use}}\mathcal{L}_{\mathrm{use}} .
	\end{aligned}
\end{equation}
The terms encourage activation sparsity, example-level evidence concentration,
selective gating, reduced redundancy, token-level coherence, and task-dependent
usage diversity or repulsion. Not all terms are active in every task. Final
weights are selected by validation trade-offs and reported in
Appendix~\ref{app:loss_weight_search}; exact definitions are given in
Appendix~\ref{app:regularizer_definitions}. When warmup or ramping is enabled,
concept-structure weights are increased gradually after an initial task-loss
phase.

\subsection{Concept support, intervention, and naming}

For classification, once the layer-attention weights have been computed for an
input, the logit is linear in the fused concept vector. We therefore use a
fixed-attention direct support score to rank layer-concept dimensions. For label
$k$, layer $\ell$, and concept dimension $j$, we define
\begin{equation}
	\mathrm{support}_{k,\ell,j}
	=
	a_{k,\ell}
	\,
	c_{\mathrm{final},j}^{(\ell)}
	\,
	w_{k,j}.
	\label{eq:concept_support}
\end{equation}
This score measures the direct contribution of a layer-concept dimension under
the layer-attention weights from the original forward pass. A concept supports
label $k$ when its activation aligns with the classifier direction for that
label. Since concept activations are signed, a negative activation combined with
a negative classifier weight can still provide positive evidence.

For qualitative classification analysis, we rank dimensions by signed support
when identifying positive evidence and by absolute support when studying total
influence. For targeted interventions, ablation is applied to the pooled
post-gate layer-level concept stack used by the prediction head. After selected
layer-concept entries are set to zero, the prediction is recomputed by passing
the ablated concept stack through the same trained layer-attention and
prediction head.

For multi-label classification, targeted ablation is label-conditioned. For each
example, we identify the positive gold labels and select the top-$k$
positive-supporting layer-concept dimensions for each such label using
Eq.~\eqref{eq:concept_support}. The intervention mask is the union of these
label-specific top-$k$ sets. If an example has no positive gold label, we use
the top predicted label as a fallback. Random ablation removes the same number
of layer-concept entries per example. This protocol tests whether the prediction
head functionally relies on the selected concept dimensions; it does not prove
that every dimension is a clean semantic concept or establish complete causal
mediation.

For Qwen verb cloze, Eq.~\eqref{eq:concept_support} does not directly apply
because the output head scores vocabulary tokens. We instead rank concept
dimensions by their gradient-times-activation contribution to the gold-token
logit and report the effect of targeted versus random ablation on the gold-token
logit, probability, and prediction flip rate.

The learned dimensions are latent variables and are not supervised with human
concept annotations. After training, we assign post-hoc names using mined
textual evidence. This naming step is not used during training or prediction.
The names are used only as diagnostic descriptors for inspection and
qualitative reporting; they are not treated as ground-truth semantic
annotations. The naming and audit protocol is described in
Section~\ref{subsec:concept_intervention_naming} and evaluated in
Appendix~\ref{app:concept_labeling_reliability}.
\section{Experimental Setup}
\label{sec:experimental_setup}

We evaluate GCA on four settings. The primary human-facing settings are
fine-grained multi-label emotion classification and multi-label toxicity
detection, which test whether a concept-mediated prediction path can support
auditable affective and safety-relevant text understanding. We also include
single-label topic classification as a compact generalization benchmark with
external concept-based baselines, and controlled full-vocabulary verb prediction
as a decoder-side boundary test. Across these settings, the goal is to evaluate
whether GCA can retain useful predictive performance while exposing
low-redundancy, layer-structured, intervention-sensitive, and auditable concept
representations. Detailed task, model, and training summaries are provided in
Appendix~\ref{app:task_model_training_summaries}.

\subsection{Datasets and task formulations}
\label{subsec:datasets_tasks}

We use fixed train/validation/test splits for all tasks. Validation data are
used for checkpoint selection, threshold tuning, and hyperparameter selection;
test data are reserved for final reporting.

\paragraph{GoEmotions}
GoEmotions is used as the main controlled ablation benchmark
\citep{demszky2020goemotions}. We use the official train/dev/test split and
treat the task as 28-label multi-label emotion classification. Because the
labels are fine-grained, imbalanced, and often overlapping, this task is used
for architecture comparisons, loss ablations, layer-routing analysis,
per-label analysis, intervention tests, and concept-name audit.

\paragraph{AG News}
AG News is used as a compact topic-classification benchmark
\citep{zhang2015character}. It contains four coarse labels: World, Sports,
Business, and Sci/Tech. We use the clean-v2 preprocessing pipeline with
tokenizer text-pair encoding, as detailed in
Appendix~\ref{app:preprocessing_details}. AG News provides the direct comparison
with external concept-based baselines, CB-LLM~\citep{sun2025concept} and
HI-Concept~\citep{zhao2024hiconcept}.

\paragraph{Toxicity detection}
For toxicity detection, we use Jigsaw/Civil Comments
\citep{borkan2019nuanced} with seven labels:
\textit{target}, \textit{severe\_toxicity}, \textit{obscene},
\textit{identity\_attack}, \textit{insult}, \textit{threat}, and
\textit{sexual\_explicit}. The \textit{severe\_toxicity} label is included
during training but excluded from aggregate validation and test metrics because
it has too few or no positive validation examples under the split used for
threshold selection. Aggregate metrics are therefore computed over six labels:
\textit{target}, \textit{obscene}, \textit{identity\_attack}, \textit{insult},
\textit{threat}, and \textit{sexual\_explicit}. We use
ModernBERT~\citep{modernbert2024} because the comments are longer and the task
is safety-relevant.

\paragraph{Qwen verb cloze}
For Qwen verb cloze, we replace a single \textsc{verb} or \textsc{aux} token in
UD English-EWT \citep{nivre2020universal,silveira2014gold} with
\texttt{[BLANK]} and predict the missing token over the full Qwen vocabulary
\citep{qwen2techreport2024,qwen25techreport2024}. We keep only examples whose
target is represented by one tokenizer token. This is a controlled single-token
prediction task, not open-ended generation.

\subsection{Model variants and training}
\label{subsec:model_variants_training}

The full architecture-control family is evaluated on GoEmotions; selected
configurations are used for AG News, toxicity detection, and Qwen verb cloze.
The main end-to-end concept model is \textsc{MLC}. We also evaluate
unconstrained fine-tuned baselines, frozen-backbone concept variants,
task-initialized frozen concept variants, adapter-only controls, and single
final-layer bottleneck controls.

The distinction among these variants is important. \textsc{MLC} tests joint
task learning and concept-interface learning. \textsc{F-MLC} tests whether a
concept readout can be learned from a frozen pretrained backbone.
\textsc{TI-F-MLC} tests a practical audit scenario in which an already
task-adapted encoder is frozen and wrapped with an auditable concept-mediated
prediction head. These variants are therefore not intended to represent a
single identical training protocol. Comparisons among them are interpreted as
accuracy--auditability and training-regime trade-offs, not as claims that all
models have identical trainable capacity.

All experiments are implemented in PyTorch with Hugging Face Transformers. We
use AdamW, gradient clipping when needed, and mixed precision when supported.
For multi-label tasks, thresholds are tuned on the validation split and fixed
for test evaluation. Classification results are reported over five random seeds
unless otherwise stated. Qwen verb cloze is reported over three seeds because
partially fine-tuning the decoder is more expensive.

Concept dimensionality, selected layers, and regularization weights are selected
using validation performance and concept diagnostics. GoEmotions receives the
most detailed loss-ablation study because it is the main controlled benchmark.
Final active loss weights, usage variants, preprocessing details, training
configurations, and runtime notes are reported in
Appendix~\ref{app:experimental_details}.

\subsection{Evaluation metrics}
\label{subsec:evaluation_metrics}

We report predictive performance and concept diagnostics separately. For
multi-label classification, we report macro-F1, micro-F1, macro-AUPRC,
micro-AUPRC, and Hamming loss where appropriate. For single-label
classification, we report accuracy, macro-F1, and log loss. For Qwen verb cloze,
we report top-1, top-5, and top-10 accuracy, mean reciprocal rank, and separate
results for \textsc{verb} and \textsc{aux} targets.

For concept diagnostics, we report top-$k$ concept mass, Gini coefficient,
entropy, active concept count, mean absolute concept correlation, layer entropy,
and layer peakiness. Redundancy is interpreted comparatively relative to matched
controls rather than as an absolute threshold of interpretability. For AG News,
we also report concept-label purity when comparing with external concept-based
baselines.

\subsection{Concept intervention and naming audit}
\label{subsec:concept_intervention_naming}

We test functional reliance through targeted concept ablation at test time. For
each example and target output, we identify high-supporting concept dimensions,
set the selected entries to zero in the pooled layer-level concept stack, and
recompute the output using the same trained layer-attention and prediction head.
For classification, concepts are ranked using the label-conditioned support
score in Eq.~\eqref{eq:concept_support}. We report targeted probability drop,
random probability drop under removal of the same number of dimensions, and
their difference:
\begin{equation}
	\Delta_{\mathrm{gap}}
	=
	\Delta_{\mathrm{targeted}}
	-
	\Delta_{\mathrm{random}} .
\end{equation}
For Qwen verb cloze, we rank dimensions by gradient-times-activation
contribution to the gold-token logit and report logit drop, probability drop,
and prediction flip rate. A large targeted-random gap indicates functional
reliance on the selected dimensions; it does not prove that every dimension is a
clean semantic concept or establish full causal mediation.

Concept names are assigned after training and are used only for inspection and
qualitative reporting. For each layer-concept dimension, we collect
high-activation $n$-grams, contrastive $n$-grams, activation-pole metadata, and
representative examples. Two instruction-tuned models propose candidate
signed-pole names, and a third instruction-tuned model acts as a referee. We
audit name grounding using three independent audit passes over 100 sampled
concept packets per audited task. Decision rules, pooled scores,
majority-consensus acceptance rates, and inter-audit agreement are reported in
Appendix~\ref{app:concept_labeling_reliability}.

\subsection{Reporting conventions}
\label{subsec:reporting_conventions}

For multi-seed experiments, we report mean and standard deviation. Single-run
external baselines are marked accordingly. Concept diagnostics and interventions
are reported only for models with concept dimensions. Per-label results for very
rare labels, such as \textit{grief} and \textit{relief} in GoEmotions, are shown
for completeness but should not be interpreted as stable estimates. All
experiments are run on an NVIDIA RTX 3090 GPU.
\section{Results}
\label{sec:results}
\begin{table}[pos=t]
	\centering
	\scriptsize
	\caption{Main predictive results on GoEmotions. Values are 5-seed test means
		$\pm$ standard deviations on the official train/dev/test split. All models use
		validation-tuned thresholds. Best values are bolded and second-best values are
		underlined.}
	\label{tab:goemotions_predictive_main}
	\begin{tabular}{lcccc}
		\toprule
		Model & Macro-F1 & Micro-F1 & Macro-AUPRC & Micro-AUPRC \\
		\midrule
		\textsc{RoBERTa}
		& \textbf{0.5230 $\pm$ 0.0108}
		& \textbf{0.6032 $\pm$ 0.0038}
		& \textbf{0.5123 $\pm$ 0.0074}
		& \underline{0.6384 $\pm$ 0.0077} \\
		
		\textsc{MLC}
		& 0.4587 $\pm$ 0.0104
		& 0.5698 $\pm$ 0.0071
		& 0.4328 $\pm$ 0.0043
		& 0.6005 $\pm$ 0.0036 \\
		
		\textsc{TI-F-MLC}
		& \underline{0.5090 $\pm$ 0.0083}
		& \underline{0.6021 $\pm$ 0.0051}
		& \underline{0.4988 $\pm$ 0.0052}
		& \textbf{0.6437 $\pm$ 0.0078} \\
		
		\textsc{F-MLC}
		& 0.3799 $\pm$ 0.0114
		& 0.5478 $\pm$ 0.0204
		& 0.3934 $\pm$ 0.0047
		& 0.6023 $\pm$ 0.0050 \\
		
		\textsc{MLA}
		& 0.4511 $\pm$ 0.0129
		& 0.5614 $\pm$ 0.0208
		& 0.4440 $\pm$ 0.0089
		& 0.5910 $\pm$ 0.0088 \\
		
		\textsc{SLC}
		& 0.4593 $\pm$ 0.0120
		& 0.5746 $\pm$ 0.0075
		& 0.4389 $\pm$ 0.0031
		& 0.5800 $\pm$ 0.0120 \\
		
		\textsc{SLA}
		& 0.4631 $\pm$ 0.0076
		& 0.5703 $\pm$ 0.0084
		& 0.4461 $\pm$ 0.0016
		& 0.5882 $\pm$ 0.0073 \\
		\bottomrule
	\end{tabular}
\end{table}
We present the results with the human-facing evaluations first. GoEmotions tests
whether GCA can expose auditable concept pathways for fine-grained affective
text prediction, and toxicity detection tests whether the same interface can
support inspection, intervention, and post-hoc control in a safety-relevant
social-text setting. AG News is used as a compact topic-classification
generalization benchmark with external concept-based baselines, while Qwen verb
cloze is used as a controlled decoder-side boundary test for full-vocabulary
prediction.

Across tasks, we interpret GCA as an accuracy--auditability trade-off. The goal
is not to universally outperform unconstrained fine-tuning, but to expose
comparatively low-redundancy, layer-structured, intervention-sensitive, and
auditable concept interfaces while retaining useful predictive performance.
\begin{table}[pos=t]
	\centering
	\scriptsize
	\caption{Compact interpretability summary on GoEmotions. Values are 5-seed
		test means $\pm$ standard deviations. Higher Top-10 Mass and Gini indicate more
		concentrated usage; lower redundancy indicates less correlated concept
		dimensions. Best and second-best values are marked among multi-layer concept
		models.}
	\label{tab:goemotions_interpretability_main}
	\begin{tabular}{lccc}
		\toprule
		Model & Top-10 Mass & Gini & Redundancy \\
		\midrule
		\textsc{RoBERTa} & -- & -- & -- \\
		\textsc{MLC}
		& 0.1330 $\pm$ 0.0030
		& 0.7454 $\pm$ 0.0046
		& 0.1784 $\pm$ 0.0042 \\
		\textsc{TI-F-MLC}
		& \textbf{0.2974 $\pm$ 0.0254}
		& \textbf{0.8956 $\pm$ 0.0079}
		& \textbf{0.1300 $\pm$ 0.0031} \\
		\textsc{F-MLC}
		& \underline{0.1764 $\pm$ 0.0249}
		& \underline{0.8297 $\pm$ 0.0235}
		& \underline{0.1508 $\pm$ 0.0092} \\
		\textsc{MLA}
		& 0.0671 $\pm$ 0.0030
		& 0.5443 $\pm$ 0.0066
		& 0.2438 $\pm$ 0.0078 \\
		\textsc{SLC}
		& 0.6198 $\pm$ 0.0358
		& 0.6641 $\pm$ 0.0254
		& 0.2037 $\pm$ 0.0099 \\
		\textsc{SLA}
		& 0.3970 $\pm$ 0.0128
		& 0.4335 $\pm$ 0.0178
		& 0.2516 $\pm$ 0.0086 \\
		\bottomrule
	\end{tabular}
\end{table}
Throughout this section, targeted-versus-random ablation is interpreted as a
test of functional reliance on the concept interface. It is not treated as proof
that every dimension is a clean semantic concept or that the full prediction is
causally mediated by human-interpretable concepts.
\begin{table}[pos=t]
	\centering
	\scriptsize
	\caption{Targeted vs. random concept ablation on GoEmotions. Drops are
		selected-label probability decreases after ablating the top-10 supporting
		concept dimensions. Values are 5-seed test means $\pm$ standard deviations. A
		larger gap indicates stronger functional reliance on the ranked dimensions; it
		is not by itself a measure of semantic interpretability.}
	\label{tab:goemotions_ablation_main}
	\begin{tabular}{lccc}
		\toprule
		Model & Targeted Drop & Random Drop & Gap \\
		\midrule
		\textsc{RoBERTa} & -- & -- & -- \\
		\textsc{MLC}
		& 0.2827 $\pm$ 0.0167
		& 0.0010 $\pm$ 0.0006
		& 0.2817 $\pm$ 0.0166 \\
		\textsc{TI-F-MLC}
		& 0.4671 $\pm$ 0.0159
		& 0.0028 $\pm$ 0.0006
		& 0.4643 $\pm$ 0.0155 \\
		\textsc{F-MLC}
		& 0.3098 $\pm$ 0.0221
		& 0.0008 $\pm$ 0.0006
		& 0.3090 $\pm$ 0.0220 \\
		\textsc{MLA}
		& 0.3706 $\pm$ 0.0224
		& 0.0030 $\pm$ 0.0003
		& 0.3676 $\pm$ 0.0225 \\
		\textsc{SLC}
		& 0.4016 $\pm$ 0.0160
		& 0.0067 $\pm$ 0.0029
		& 0.3949 $\pm$ 0.0181 \\
		\textsc{SLA}
		& 0.4876 $\pm$ 0.0218
		& 0.0083 $\pm$ 0.0028
		& 0.4793 $\pm$ 0.0203 \\
		\bottomrule
	\end{tabular}
\end{table}

\subsection{Human-Facing Affective Prediction with RoBERTa}
\label{subsec:results_goemotions}

GoEmotions is the main controlled benchmark because it combines multi-label
supervision, fine-grained affective distinctions, label imbalance, and semantic
overlap among classes. We compare the task-finetuned RoBERTa baseline with the
selected end-to-end multi-layer concept model (\textsc{MLC}), a
task-initialized frozen-backbone variant (\textsc{TI-F-MLC}), a
frozen-pretrained-backbone variant (\textsc{F-MLC}), a multi-layer adapter-only
control (\textsc{MLA}), and single final-layer controls with and without concept
regularization (\textsc{SLC} and \textsc{SLA}). All multi-label results in this
section use validation-tuned thresholds selected on the development split and
then fixed for test evaluation. The corresponding loss weights are reported in
Table~\ref{tab:selected_loss_weights_all_tasks}, and detailed loss ablations are
reported in Appendix~\ref{app:goemotions_loss_ablation}.

\subsubsection{Predictive performance}
\label{subsubsec:goemotions_predictive}

Table~\ref{tab:goemotions_predictive_main} reports the main predictive results.
The unconstrained RoBERTa baseline remains the strongest model for raw macro-
and micro-level prediction. This is expected because it can use the full
fine-tuned hidden representation directly, whereas \textsc{MLC} constrains the
prediction path through a concept interface.

The end-to-end \textsc{MLC} should therefore not be read as an accuracy
improvement over RoBERTa. Its role is to test how much predictive utility can be
retained while exposing layer-wise concept summaries that can be inspected and
intervened on. \textsc{TI-F-MLC} gives the strongest concept-bearing predictive
result, but it represents a different training regime: it starts from a
task-adapted frozen encoder and learns an auditable concept readout on top.
Thus, \textsc{TI-F-MLC} should be interpreted as a practical conversion or audit
setting rather than as evidence that end-to-end concept learning is cost-free.

A label-wise analysis shows that the end-to-end concept bottleneck does not
affect all emotions equally. The performance cost is concentrated most strongly
on rare labels, where the unconstrained fine-tuned encoder has an advantage.
For labels with more support, the gap is smaller, and \textsc{TI-F-MLC}
substantially reduces it. Detailed support-bin and representative label-wise
results are reported in Appendix~\ref{app:goemotions_labelwise_gap}.

\subsubsection{Concept structure and intervention}
\label{subsubsec:goemotions_structure_intervention}

Table~\ref{tab:goemotions_interpretability_main} summarizes concept
concentration and redundancy. These diagnostics are interpreted comparatively,
relative to matched controls trained on the same task, rather than as absolute
thresholds of interpretability.

The key internal control is the comparison between \textsc{MLC} and
\textsc{MLA}. Both models use a multi-layer bottleneck-style architecture, but
\textsc{MLA} is trained only with task loss. It remains predictively useful, yet
its concept usage is more diffuse and more redundant. In contrast,
\textsc{MLC} produces a more concentrated and less redundant concept space,
indicating that the concept-oriented objective contributes structure beyond
adapter capacity.

\textsc{TI-F-MLC} combines strong predictive performance with the most
concentrated and least redundant multi-layer concept profile, making it a
useful diagnostic variant over a task-adapted encoder. The single-layer models
should be interpreted separately: their concept mass can be highly concentrated
because all prediction-facing evidence is forced through one final-layer
bottleneck, but this does not provide the same layer-wise routing information as
the multi-layer concept interface.

Table~\ref{tab:goemotions_ablation_main} reports targeted-versus-random concept
ablation. Across all concept-bearing models, removing the highest-supporting
dimensions changes the selected-label probability much more than removing random
dimensions of the same size. For the selected \textsc{MLC}, the random drop is
almost zero, while targeted removal produces a clear probability decrease. The
same pattern becomes stronger as more top-ranked dimensions are removed, as
shown in Appendix~\ref{app:goemotions_topk_ablation}.

The ablation gap should not be interpreted alone as semantic interpretability.
For example, \textsc{SLA} has the largest gap, but it is a single-bottleneck
adapter-only model, so targeted removal is naturally highly disruptive. The
most reliable reading comes from combining intervention sensitivity with
concentration, redundancy, layer routing, and the post-hoc naming audit.

\subsubsection{Layer-wise routing and loss ablations}
\label{subsubsec:goemotions_layer_loss}

\begin{figure}[pos=t]
	\centering
	\includegraphics[width=0.48\linewidth]{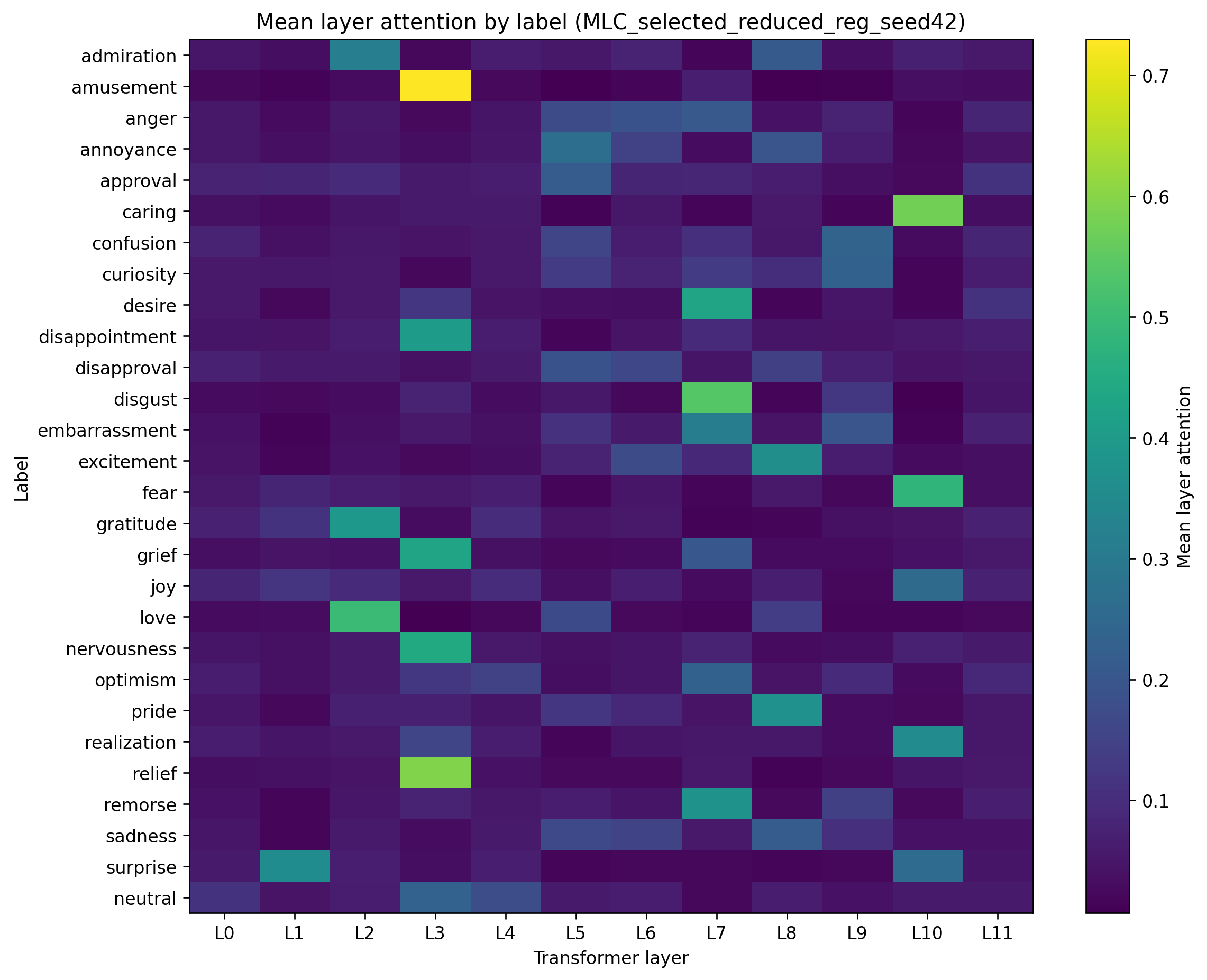}
	\includegraphics[width=0.48\linewidth]{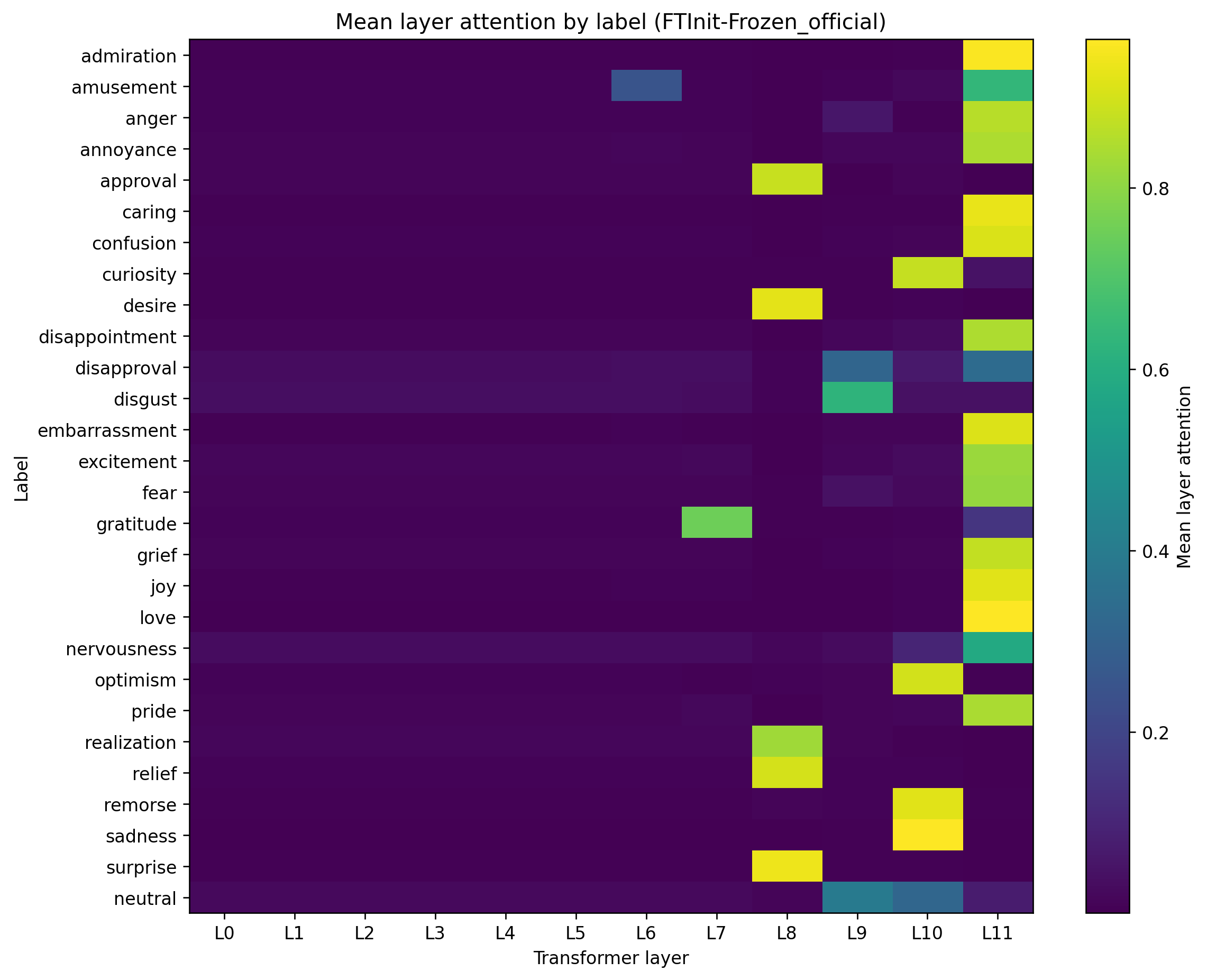}
	\caption{Mean label-wise layer attention on GoEmotions. Left: selected
		\textsc{MLC}. Right: \textsc{TI-F-MLC}. Rows are emotion labels and columns are
		Transformer layers. Attention here refers to the learned label-wise
		layer-attention head over concept summaries, not Transformer self-attention.}
	\label{fig:goemotions_layer_attention}
\end{figure}

Figure~\ref{fig:goemotions_layer_attention} analyzes the learned label-wise
layer-attention head over concept summaries. This is not Transformer
self-attention; it shows which layer-level concept vectors the prediction head
uses for each emotion label.

The two panels reveal different but coherent routing behavior. The end-to-end
\textsc{MLC} uses a more distributed profile, with labels peaking at different
early, middle, and late layers. This supports the motivation for a layer-wise
concept interface rather than a single final-layer bottleneck. The
task-initialized frozen variant is sharper and more late-layer dominated, which
is expected because the upper layers of the encoder have already been
task-adapted before the concept readout is trained.

The detailed loss ablations are reported in
Appendix~\ref{app:goemotions_loss_ablation}. They show why the selected
configuration uses reduced-strength concept regularization and disables usage
diversity. In this fine-grained multi-label setting, affective evidence is often
localized and shared across labels; stronger usage-diversity pressure can
over-constrain the representation, while overly strong sparsity can hurt
prediction.

Representative GoEmotions concept evidence and example-level interventions are
reported in Appendix~\ref{app:goemotions_qualitative}. These examples are
illustrative rather than standalone proof of interpretability; the aggregate
evidence is provided by the diagnostics and intervention results above.

\subsection{Topic Classification as a Generalization Check}
\label{subsec:results_agnews}

AG News provides the direct comparison with external concept-based baselines.
Unlike GoEmotions, it has only four coarse topic labels and clearer separation
among classes. It is therefore a useful setting for evaluating whether GCA can
recover compact, class-relevant evidence in a simpler classification regime.

\subsubsection{Predictive performance}
\label{subsubsec:agnews_predictive}

Table~\ref{tab:agnews_predictive} reports the AG News predictive results. The
strongest systems are close to one another, as expected for this relatively
clean topic-classification benchmark. \textsc{TI-F-MLC} obtains the highest
mean accuracy and macro-F1, while \textsc{MLC} remains in the same performance
range as RoBERTa, CB-LLM, and HI-Concept. CB-LLM obtains the lowest log loss in
the reproduced single run.

We interpret this as a positive but bounded result. AG News shows that the
layer-wise concept interface can preserve strong predictive performance in a
directly comparable concept-benchmark setting. It does not imply that GCA is
universally more accurate than unconstrained fine-tuning or all prior concept
methods.

\begin{table}[pos=t]
	\centering
	\scriptsize
	\caption{AG News predictive performance. Values are mean $\pm$ standard
		deviation over five seeds when available. CB-LLM is reported from a single
		reproduced run.}
	\label{tab:agnews_predictive}
	\begin{tabular}{lccc}
		\toprule
		Model & Accuracy $\uparrow$ & Macro-F1 $\uparrow$ & Log loss $\downarrow$ \\
		\midrule
		\textsc{RoBERTa} & $0.9484 \pm 0.0023$ & $0.9485 \pm 0.0023$ & $0.1927 \pm 0.0170$ \\
		\textsc{MLA} & $0.9407 \pm 0.0011$ & $0.9407 \pm 0.0010$ & $0.2110 \pm 0.0236$ \\
		\textsc{MLC} & $0.9441 \pm 0.0023$ & $0.9440 \pm 0.0024$ & $0.2278 \pm 0.0131$ \\
		\textsc{TI-F-MLC} & $\mathbf{0.9508 \pm 0.0010}$ & $\mathbf{0.9509 \pm 0.0010}$ & $0.1912 \pm 0.0158$ \\
		CB-LLM & $0.9453$ & $0.9453$ & $\mathbf{0.1824}$ \\
		HI-Concept & $0.9469 \pm 0.0003$ & $0.9468 \pm 0.0003$ & $0.1860 \pm 0.0008$ \\
		\bottomrule
	\end{tabular}
\end{table}

\subsubsection{Concept diagnostics and intervention}
\label{subsubsec:agnews_interpretability}

Table~\ref{tab:agnews_interpretability} compares the concept spaces and
intervention behavior. The adapter-only control, \textsc{MLA}, is the key
internal baseline: it uses a similar multi-layer adaptation structure but is
trained only for prediction. Its weak concentration, higher redundancy, and
small ablation gap show that architecture alone does not create a useful concept
interface.

The GCA variants improve the main structural and intervention-oriented
diagnostics relative to this adapter-only control: they produce more
concentrated concept usage, lower redundancy, and substantially stronger
targeted-versus-random intervention effects. Concept-label purity shows a more
mixed pattern. \textsc{MLC} modestly improves purity relative to \textsc{MLA},
whereas \textsc{TI-F-MLC} prioritizes predictive performance, low redundancy,
and intervention sensitivity rather than class-purity alignment.

The external baselines highlight different trade-offs. CB-LLM uses externally
generated class-conditioned concepts, while HI-Concept is a high-impact-concept
method with very concentrated but more redundant concept usage. GCA occupies a
different point in this space: its concepts are learned internally, named only
after training, and remain comparatively low-redundancy while still being
intervention-sensitive.

\begin{table}[pos=t]
	\centering
	\scriptsize
	\caption{AG News concept diagnostics and intervention results. The ablation gap
		is the targeted-minus-random drop in predicted-class probability. A larger gap
		indicates stronger functional reliance on the ranked concept dimensions; lower
		redundancy and random drop are better.}
	\label{tab:agnews_interpretability}
	\begin{tabular}{lccccc}
		\toprule
		Model & Top-10 mass $\uparrow$ & Redund. $\downarrow$ & Purity $\uparrow$ & Abl. gap $\uparrow$ & Rand. drop $\downarrow$ \\
		\midrule
		\textsc{MLA} & $0.103 \pm 0.002$ & $0.453 \pm 0.011$ & $0.470 \pm 0.005$ & $0.049 \pm 0.017$ & $0.0010 \pm 0.0002$ \\
		\textsc{MLC} & $0.499 \pm 0.029$ & $0.204 \pm 0.009$ & $0.510 \pm 0.015$ & $0.395 \pm 0.031$ & $0.0012 \pm 0.0002$ \\
		\textsc{TI-F-MLC} & $0.460 \pm 0.047$ & $\mathbf{0.178 \pm 0.003}$ & $0.385 \pm 0.023$ & $0.426 \pm 0.059$ & $0.0012 \pm 0.0003$ \\
		CB-LLM & $0.321$ & $0.275$ & $\mathbf{0.820}$ & $0.309$ & $0.0036$ \\
		HI-Concept & $\mathbf{0.998 \pm 0.000}$ & $0.671 \pm 0.014$ & $0.427 \pm 0.013$ & $\mathbf{0.830 \pm 0.057}$ & $0.0269 \pm 0.0034$ \\
		\bottomrule
	\end{tabular}
\end{table}

Representative AG News qualitative examples are summarized in
Appendix~\ref{app:agnews_qualitative}. They show that high-support concept
packets recover recognizable topic evidence such as geopolitical conflict for
World, teams and scores for Sports, profit and oil-price cues for Business, and
software, security, and technology entities for Sci/Tech.

\subsection{Safety-Relevant Toxicity Classification with ModernBERT}
\label{subsec:results_toxicity}

We evaluate toxicity detection as a safety-relevant multi-label setting where
auditability is especially important. The model is trained with seven outputs:
\textit{target}, \textit{severe\_toxicity}, \textit{obscene},
\textit{identity\_attack}, \textit{insult}, \textit{threat}, and
\textit{sexual\_explicit}. Because \textit{severe\_toxicity} has too few or no
positive validation examples under the split used for threshold selection, it
is excluded from aggregate valid-label metrics.

\subsubsection{Predictive performance}
\label{subsubsec:toxicity_predictive}

Table~\ref{tab:toxicity_predictive_compact} compares
\textsc{ModernBERT-MLC} with a plain fine-tuned
\textsc{FT-ModernBERT-MLP} baseline under the same data split, label space,
threshold-tuning protocol, and reporting convention. The unconstrained baseline
remains the strongest predictive model across the reported metrics. The
important result is that the predictive cost is moderate on the primary
\textit{target} label: \textsc{ModernBERT-MLC} remains close in target F1,
target AUC, and target AUPRC while exposing concept dimensions that can be
inspected and intervened on.

\begin{table}[pos=t]
	\centering
	\scriptsize
	\caption{Predictive performance on toxicity detection. Values are 5-seed test
		means $\pm$ standard deviations. Valid-label metrics exclude
		\textit{severe\_toxicity}. \textsc{FT-MBERT} denotes
		\textsc{FT-ModernBERT-MLP}.}
	\label{tab:toxicity_predictive_compact}
	\begin{tabular}{lcc}
		\toprule
		Metric & \textsc{FT-MBERT} & \textsc{ModernBERT-MLC} \\
		\midrule
		Target F1 & $\mathbf{0.7156 \pm 0.0009}$ & $0.7078 \pm 0.0039$ \\
		Target AUC & $\mathbf{0.9731 \pm 0.0001}$ & $0.9712 \pm 0.0008$ \\
		Target AUPRC & $\mathbf{0.8085 \pm 0.0003}$ & $0.7978 \pm 0.0050$ \\
		Valid-label Macro-F1 & $\mathbf{0.5942 \pm 0.0017}$ & $0.5785 \pm 0.0066$ \\
		Valid-label Micro-F1 & $\mathbf{0.6957 \pm 0.0004}$ & $0.6864 \pm 0.0036$ \\
		Valid-label Macro-AUPRC & $\mathbf{0.6386 \pm 0.0016}$ & $0.6095 \pm 0.0109$ \\
		Valid-label Micro-AUPRC & $\mathbf{0.5861 \pm 0.0012}$ & $0.5738 \pm 0.0049$ \\
		Valid-label Hamming Loss $\downarrow$ & $\mathbf{0.0162 \pm 0.0001}$ & $0.0169 \pm 0.0003$ \\
		\bottomrule
	\end{tabular}
\end{table}

\subsubsection{Concept structure and intervention sensitivity}
\label{subsubsec:toxicity_interpretability_ablation}

Table~\ref{tab:toxicity_concept_intervention_compact} summarizes the learned
concept space and intervention results for \textsc{ModernBERT-MLC}. The concept
interface is highly concentrated: most concept mass is carried by a small set of
dimensions, the average number of active concepts is small relative to the full
inventory, and mean absolute concept correlation remains low. This indicates
that the model does not simply spread toxicity evidence across many redundant
dimensions.

The intervention results show that these concentrated dimensions are
functionally used. When the highest-supporting dimensions are removed on
true-positive examples, the predicted toxicity probability drops substantially.
Random removal of the same number of dimensions has almost no effect. The same
pattern holds when the intervention is averaged over the valid labels.

\begin{table}[pos=t]
	\centering
	\scriptsize
	\caption{Concept diagnostics and intervention results for
		\textsc{ModernBERT-MLC} on toxicity detection. Values are 5-seed test means
		$\pm$ standard deviations. Ablation drops are probability decreases on
		true-positive examples.}
	\label{tab:toxicity_concept_intervention_compact}
	\begin{tabular}{llc}
		\toprule
		Group & Metric & Value \\
		\midrule
		Concept space & Entropy & $1.6364 \pm 0.1852$ \\
		Concept space & Top-10 mass & $0.9594 \pm 0.0064$ \\
		Concept space & Gini & $0.9937 \pm 0.0006$ \\
		Concept space & Active concepts $>0.10$ & $22.09 \pm 3.64$ \\
		Concept space & Redundancy & $0.0542 \pm 0.0161$ \\
		Concept space & Layer entropy / peakiness & $1.8124 \pm 0.1992$ / $0.5616 \pm 0.0491$ \\
		\midrule
		Intervention & \textit{target}: targeted / random / gap
		& $0.3471 \pm 0.0493$ / $0.0030 \pm 0.0003$ / $0.3441 \pm 0.0491$ \\
		Intervention & Valid-label macro: targeted / random / gap
		& $0.2347 \pm 0.0185$ / $0.0001 \pm 0.0005$ / $0.2346 \pm 0.0182$ \\
		\bottomrule
	\end{tabular}
\end{table}

\subsubsection{Post-hoc concept-level debiasing}
\label{subsubsec:toxicity_concept_debiasing}

A concept bottleneck is not automatically less biased. Identity mentions can
still be encoded in learned dimensions because they are part of the training
distribution. We therefore test whether the learned concept representation can
serve as a compact surface for post-hoc control.

Starting from a trained \textsc{ModernBERT-MLC} checkpoint, we freeze the
ModernBERT encoder, concept adapters, layer-attention module, and classifier
weights. We then learn only a small label-specific multiplicative gate over the
fused concept representation, together with a regularized label-bias offset:
\begin{equation}
	o^{\mathrm{debias}}_k
	=
	\sum_c
	\tilde{c}_{k,c}
	W_{k,c}
	g_{k,c}
	+
	b_k
	+
	\Delta b_k ,
	\label{eq:toxicity_debias_gate}
\end{equation}
where $\tilde{c}_{k,c}$ is the fused concept value for label $k$, $W_{k,c}$ and
$b_k$ are frozen classifier parameters, $g_{k,c}$ is the learned post-hoc gate,
and $\Delta b_k$ is a small learned bias correction. The gate is initialized
near the identity mapping and is regularized to remain close to it. No manual
concept names, manually selected identity concepts, or encoder retraining are
used.

The gate is trained on cached concept features with a validation-selected
objective that balances preserving the original toxicity predictor, reducing
identity-related false positives, and reducing sensitivity to counterfactual
identity substitutions. Counterfactual sensitivity is measured by the absolute
change in the target logit:
\begin{equation}
	|\Delta z|
	=
	\left|
	z_{\mathrm{target}}(x_{\mathrm{orig}})
	-
	z_{\mathrm{target}}(x_{\mathrm{cf}})
	\right| .
	\label{eq:toxicity_cf_delta_logit}
\end{equation}

We report subgroup AUC, BPSN AUC, and BNSP AUC to summarize subgroup-related
ranking behavior, together with counterfactual logit shifts, counterfactual
consistency, expected calibration error (ECE), and Brier score.
Table~\ref{tab:toxicity_concept_debiasing} shows a clear
robustness--performance trade-off. The post-hoc gate substantially improves BPSN
AUC, reduces the worst-group accuracy gap, and nearly halves both the mean and
high-percentile counterfactual logit shift. Counterfactual consistency also
improves. However, target F1, target AUPRC, and BNSP AUC decrease, and
calibration is mixed. Thus, \textsc{ModernBERT-MLC-Debias} should not be read
as uniformly better or as eliminating bias. The more specific result is that the
learned concept representation can be used after training as a lightweight,
auditable control surface for reducing identity-mention sensitivity.

\begin{table}[pos=t]
	\centering
	\scriptsize
	\caption{Post-hoc concept-level debiasing on toxicity detection. The encoder,
		concept adapters, layer-attention module, and classifier weights are frozen;
		only a label-specific gate over the fused concept representation and a
		regularized label-bias offset are learned. Higher is better except Gap, CF
		$|\Delta z|$, ECE, and Brier.}
	\label{tab:toxicity_concept_debiasing}
	\begin{tabular}{lcc}
		\toprule
		Metric & \textsc{ModernBERT-MLC} & \textsc{ModernBERT-MLC-Debias} \\
		\midrule
		Target F1 & $\mathbf{0.7077 \pm 0.0038}$ & $0.6983 \pm 0.0045$ \\
		Target AUC & $\mathbf{0.9712 \pm 0.0008}$ & $0.9693 \pm 0.0013$ \\
		Target AUPRC & $\mathbf{0.7978 \pm 0.0050}$ & $0.7839 \pm 0.0058$ \\
		Subgroup AUC & $\mathbf{0.8953 \pm 0.0027}$ & $0.8924 \pm 0.0025$ \\
		BPSN AUC & $0.9045 \pm 0.0024$ & $\mathbf{0.9344 \pm 0.0053}$ \\
		BNSP AUC & $\mathbf{0.9696 \pm 0.0009}$ & $0.9512 \pm 0.0057$ \\
		Worst-group acc. gap $\downarrow$ & $0.1826 \pm 0.0194$ & $\mathbf{0.1600 \pm 0.0074}$ \\
		CF mean $|\Delta z|$ $\downarrow$ & $0.4479 \pm 0.0176$ & $\mathbf{0.2355 \pm 0.0177}$ \\
		CF P90 $|\Delta z|$ $\downarrow$ & $1.0959 \pm 0.0451$ & $\mathbf{0.5865 \pm 0.0476}$ \\
		CF consistency & $0.9291 \pm 0.0110$ & $\mathbf{0.9595 \pm 0.0093}$ \\
		Target ECE $\downarrow$ & $\mathbf{0.2786 \pm 0.0025}$ & $0.3152 \pm 0.0099$ \\
		Target Brier $\downarrow$ & $0.1520 \pm 0.0014$ & $\mathbf{0.1498 \pm 0.0059}$ \\
		\bottomrule
	\end{tabular}
\end{table}

Representative toxicity concept-level examples are provided in
Appendix~\ref{app:toxicity_qualitative}. They illustrate direct insult
detection, false positives from strongly dismissive rhetoric, and remaining
difficulty with identity/topic framing.
\begin{table}[pos=t]
	\centering
	\scriptsize
	\caption{Qwen verb-cloze predictive performance over three seeds. The task
		predicts a masked \textsc{verb}/\textsc{aux} token over the full Qwen
		vocabulary. Values are mean $\pm$ standard deviation.}
	\label{tab:qwen_verb_performance}
	\begin{tabular}{lcc}
		\toprule
		Metric & Qwen-FT-NoConcept & Qwen-\textsc{MLC} \\
		\midrule
		Top-1 accuracy & \textbf{0.5271 $\pm$ 0.0009} & 0.5077 $\pm$ 0.0100 \\
		Top-5 accuracy & \textbf{0.8100 $\pm$ 0.0036} & 0.7566 $\pm$ 0.0107 \\
		Top-10 accuracy & \textbf{0.8676 $\pm$ 0.0013} & 0.8088 $\pm$ 0.0090 \\
		MRR & \textbf{0.6559 $\pm$ 0.0014} & 0.6231 $\pm$ 0.0099 \\
		\midrule
		Top-1 on \textsc{aux} & 0.6431 $\pm$ 0.0030 & \textbf{0.6560 $\pm$ 0.0131} \\
		Top-1 on \textsc{verb} & \textbf{0.4336 $\pm$ 0.0008} & 0.3882 $\pm$ 0.0077 \\
		\bottomrule
	\end{tabular}
\end{table}
\begin{table}[pos=t]
	\centering
	\scriptsize
	\caption{Qwen-\textsc{MLC} concept diagnostics and intervention results over
		three seeds.}
	\label{tab:qwen_verb_concepts}
	\begin{tabular}{lc}
		\toprule
		Metric & Value \\
		\midrule
		Concept dimensions & $12 \times 64 = 768$ \\
		Per-example top-10 concept mass & $0.1661 \pm 0.0091$ \\
		Per-example Gini & $0.7576 \pm 0.0128$ \\
		Active dimensions $>0.1$ & $546.4 \pm 39.6$ / 768 \\
		Mean absolute concept correlation & $0.1026 \pm 0.0044$ \\
		Final-layer usage mass & $33.2\% \pm 4.5\%$ \\
		Last-three-layer usage mass & $68.2\% \pm 1.8\%$ \\
		\midrule
		Targeted / random gold-logit drop & $5.4013 \pm 0.1581$ / $0.1600 \pm 0.0029$ \\
		Targeted / random gold-probability drop & $0.3160 \pm 0.0136$ / $0.0040 \pm 0.0002$ \\
		Targeted / random prediction flip rate & $36.1\% \pm 0.8\%$ / $3.0\% \pm 0.2\%$ \\
		\bottomrule
	\end{tabular}
\end{table}
\subsection{Controlled Verb-Cloze Prediction as a Boundary Test}
\label{subsec:results_qwen}

We use Qwen verb cloze as a controlled decoder-side stress test rather than as a
standard language-modeling benchmark. Both Qwen-\textsc{MLC} and the matched
Qwen-FT-NoConcept baseline use the same checkpoint, prompt format, data splits,
single-token target filtering, and partial fine-tuning policy. The task is
deliberately narrow: a single \textsc{verb} or \textsc{aux} token is replaced
with \texttt{[BLANK]}, and the model predicts the missing token over the full
Qwen vocabulary.

\subsubsection{Task performance}
\label{subsubsec:qwen_task_performance}

Table~\ref{tab:qwen_verb_performance} compares Qwen-\textsc{MLC} with the
matched no-concept baseline. The no-concept baseline remains stronger overall,
showing a real cost of forcing full-vocabulary prediction through a
lower-dimensional concept-mediated path. The auxiliary--lexical split is more
nuanced: Qwen-\textsc{MLC} obtains a higher mean top-1 score on \textsc{aux}
targets, whereas the no-concept baseline remains clearly better on lexical
\textsc{verb} targets. We interpret this pattern cautiously. The concept
interface appears to preserve some local syntactic-frame information, while
exact lexical recovery remains harder.

\subsubsection{Concept diagnostics and intervention}
\label{subsubsec:qwen_concept_analysis}

Table~\ref{tab:qwen_verb_concepts} reports concept diagnostics and intervention
results for Qwen-\textsc{MLC}. The representation is not sparse in the same way
as the classification models: many dimensions are active and the top-10 concept
mass is relatively small. This is expected because full-vocabulary prediction is
a broader output problem than choosing among a small set of labels.

The useful finding is that the concept path is still structured and
functionally used. Usage is concentrated in the last decoder layers, and
targeted interventions have a large effect on the gold-token score, while random
interventions of the same size have little effect. Thus, the Qwen experiment
should be read as a boundary test: it does not show improved language modeling
or highly sparse semantic decomposition, but it does show that a decoder-side
concept-mediated path can be attached to full-vocabulary scoring and still
expose intervention-sensitive internal structure.

Representative Qwen verb-cloze examples are provided in
Appendix~\ref{app:qwen_qualitative}. They illustrate the same pattern as the
aggregate results: concept-mediated prediction often captures the local
grammatical frame, but exact lexical selection remains harder.
\section{Discussion and Limitations}
\label{sec:discussion}

The results support a cautious but useful conclusion: GCA is not a universal
replacement for unconstrained fine-tuning, but it provides a prediction-path
concept interface that makes Transformer decisions easier to inspect, diagnose,
and intervene on. Its value is therefore not only predictive performance, but
also the ability to expose layer-wise concept summaries, label-conditioned
support, post-hoc concept names, and targeted intervention behavior.

\subsection{Accuracy--auditability trade-off}

Across the evaluated tasks, unconstrained fine-tuning remains a strong choice
when raw predictive performance is the only objective. This is clearest on
GoEmotions, where the end-to-end \textsc{MLC} pays a visible macro-F1 cost,
especially on rare labels, and on Qwen verb cloze, where the matched
no-concept decoder baseline remains stronger for full-vocabulary ranking. These
results show that routing predictions through a concept-mediated interface is a
real constraint rather than a cost-free modification.

The benefit of this constraint is auditability. On GoEmotions, the
adapter-only control remains predictively useful but produces a more diffuse
and redundant concept space, indicating that a bottleneck-style architecture
alone is not sufficient. Concept-regularized GCA variants produce more
concentrated, comparatively lower-redundancy, and intervention-sensitive
representations. The task-initialized frozen variant, \textsc{TI-F-MLC}, shows
a complementary diagnostic regime: a task-adapted encoder can be frozen and
wrapped with an auditable concept readout while greatly reducing the predictive
gap. This result is practically useful, but it should not be interpreted as
evidence that end-to-end concept learning is cost-free or equivalent to standard
fine-tuning.

The trade-off varies by task. On AG News, where labels are coarse and
topically separated, GCA remains competitive with recent concept-based baselines
while exposing internally learned layer-wise concepts. On toxicity detection,
\textsc{ModernBERT-MLC} remains close to the unconstrained baseline on the main
toxicity target while supporting concept inspection, targeted intervention, and
lightweight post-hoc control. On Qwen verb cloze, the concept path is more
distributed and less predictive than the no-concept baseline, but it still
reveals late-layer reliance and strong targeted-versus-random intervention
effects. Overall, GCA is most appropriate when inspection, diagnosis, or
intervention is important enough to justify some loss of unconstrained
predictive flexibility.

\subsection{Layer-wise concepts, regularization, and interventions}

The layer-wise design is useful because different labels need not rely on the
same encoder depth. The GoEmotions routing analysis shows that the end-to-end
\textsc{MLC} uses a distributed label-specific profile, while
\textsc{TI-F-MLC} is sharper and more late-layer dominated. This distinction is
meaningful: the end-to-end model collects prediction-facing evidence from
several abstraction levels, whereas the task-initialized frozen readout can
rely more heavily on upper layers that have already become task-adapted.

The ablations also show that concept structure should not be assumed to emerge
from task loss alone. A model can learn useful latent features while still
producing dense, redundant, or difficult-to-audit dimensions. The
concept-structure losses are therefore best understood as modular pressures
against specific failure modes: diffuse activation, duplicated dimensions,
incoherent token evidence, always-open gates, and unhelpful usage collapse.
They are not a fixed universal recipe. Their weights, and even whether a term is
active, depend on the task and are selected through validation trade-offs.

Targeted-versus-random ablation provides the main functional test of the
concept interface. Static diagnostics can show whether a concept space is
concentrated or redundant, but they do not show whether the prediction head
actually uses the exposed dimensions. Across tasks, targeted removal changes
outputs much more than random removal of the same size, indicating that the
selected dimensions are operationally relevant. This result should still be
interpreted carefully: a large targeted-random gap measures reliance inside the
learned interface; it does not prove that every dimension is a clean semantic
concept or that the full model behavior is causally mediated by
human-interpretable concepts.

\subsection{Post-hoc names and post-hoc control}

Post-hoc concept names serve a different purpose from interventions. They are
human-readable descriptors of mined evidence, not training labels or definitive
semantic annotations. The LLM-assisted labeling and Qwen-referee pipeline makes
the learned dimensions easier to inspect, but some names remain broad,
partially grounded, attached to the wrong activation pole, or occasionally
misleading. For this reason, we evaluate name grounding separately in
Appendix~\ref{app:concept_labeling_reliability}. The audit asks whether the
displayed textual evidence supports the assigned name, while the intervention
experiments ask whether the underlying dimension is used by the prediction head.

The toxicity experiment shows an additional use of the concept interface as a
post-hoc control surface. The debiasing variant does not retrain the encoder or
manually select identity concepts; it learns only a small concept-level gate and
regularized bias offset on top of the frozen concept representation. The result
is a clear robustness--performance trade-off: identity-mention false-positive
behavior and counterfactual sensitivity improve, while some predictive and
calibration metrics degrade. This is not a complete fairness solution, and it
does not eliminate bias. Rather, it demonstrates that exposing a compact
concept-mediated interface can make some post-hoc behavioral control possible
without retraining the full Transformer backbone.

\subsection{Limitations}

GCA has several limitations. First, learned dimensions are not guaranteed to be
perfectly human-nameable. Some dimensions capture mixed lexical, syntactic,
topical, affective, or dataset-specific patterns, especially in short and
multi-label text. Concept names should therefore be read as evidence-grounded
descriptors rather than as ground-truth semantic concepts.

Second, GCA introduces additional design choices. Concept dimensionality,
selected layers, regularization weights, gate behavior, pooling design, and
usage terms affect the accuracy--auditability trade-off. We select these
settings through validation sweeps and ablations, but broader capacity studies
over more backbones, concept dimensions, and task families remain future work.

Third, the GCA modules are prediction-facing readout adapters rather than
sequential bottlenecks that replace the Transformer hidden stream. This design
preserves the expressive pretrained backbone and makes the task head
concept-mediated, but it does not force every intermediate computation inside
the Transformer to pass through concepts. Thus, GCA provides an auditable
prediction path, not a complete mechanistic decomposition of the entire
backbone.

Fourth, external concept-baseline comparison is limited to AG News, where direct
task overlap is available. Extending those comparisons to fine-grained emotion
classification, toxicity detection, and controlled token prediction would
require adapting or reimplementing prior methods under different task
assumptions.

Fifth, the Qwen experiment is deliberately narrow. It is a controlled
single-token verb-cloze stress test, not open-ended generation. It shows that a
decoder-side concept-mediated path can be connected to full-vocabulary scoring
and intervened on, but it does not show that GCA transfers unchanged to
long-form generation, dialogue, instruction following, or multi-token decoding.

Finally, GCA adds computational and analysis cost. Extracting layer-wise
concepts, mining evidence, running interventions, assigning names, and auditing
those names require work beyond standard fine-tuning. These costs are most
justified when auditability, diagnosis, or post-hoc intervention is part of the
modeling goal.

\section{Conclusion}
\label{sec:conclusion}

We introduced Gated Concept Adapters, a layer-wise concept bottleneck framework
for pretrained Transformers. GCA attaches gated concept-producing readout
adapters to selected Transformer layer outputs, pools token-level activations
into layer-level concept summaries, and uses those summaries as the
prediction-facing interface. This design makes predictions inspectable through
concept usage, layer-wise routing, post-hoc names, and targeted interventions.

Across GoEmotions, AG News, toxicity detection, and controlled Qwen verb cloze,
GCA exposes structured and intervention-sensitive concept pathways while
retaining useful predictive performance in classification settings. The results
also clarify the limits of the approach: concept bottlenecks do not
automatically improve raw accuracy, post-hoc names are not ground-truth semantic
annotations, and decoder-side full-vocabulary prediction remains harder than
fixed-label classification. The strongest supported claim is therefore not
universal superiority, but useful auditability. GCA provides a prediction-path
concept interface that can make Transformer decisions easier to inspect,
diagnose, and control.

These results suggest that concept-mediated neural interfaces are especially
useful when human-facing affective or safety-sensitive systems require
inspection, diagnosis, and post-hoc control. Future work should extend GCA to
multi-token generation, larger decoder-only and encoder--decoder models,
broader human audits, and interactive settings where users can edit or suppress
concept pathways directly. These directions would test whether layer-wise
concept interfaces can support not only post-hoc explanation, but also practical
human-in-the-loop control of language models.
\appendix

\counterwithin{table}{section}
\counterwithin{figure}{section}
\counterwithin{equation}{section}

\renewcommand{\thetable}{\Alph{section}.\arabic{table}}
\renewcommand{\thefigure}{\Alph{section}.\arabic{figure}}
\renewcommand{\theequation}{\Alph{section}.\arabic{equation}}

\FloatBarrier
\clearpage
\section{Regularizer Definitions}
\label{app:regularizer_definitions}

This appendix defines the concept-structure losses used in GCA. The losses are
modular: each term targets a specific failure mode of latent concept learning,
and not all terms are active in every experiment. The selected task-specific
weights are reported in Appendix~\ref{app:loss_weight_search}.

\begin{table}[pos=t]
	\centering
	\footnotesize
	\caption{Role of the concept regularizers.}
	\label{tab:loss_intuition}
	\begin{tabularx}{\linewidth}{p{0.16\linewidth} X X}
		\toprule
		Term & Encouraged behavior & Failure mode addressed \\
		\midrule
		$\mathcal{L}_{1}$ &
		Only a few concepts should have large values for an example. &
		Dense explanations with many weakly active dimensions. \\
		
		$\mathcal{L}_{\mathrm{excl}}$ &
		Evidence should concentrate in the most relevant dimensions. &
		The same prediction is spread thinly across many small activations. \\
		
		$\mathcal{L}_{\mathrm{gate}}$ &
		Token-level concept channels should open selectively. &
		All channels remain open for all tokens. \\
		
		$\mathcal{L}_{\mathrm{orth}}$ &
		Different dimensions should encode distinct evidence. &
		Several concept dimensions duplicate the same cue. \\
		
		$\mathcal{L}_{\mathrm{intra}}$ &
		Tokens assigned to the same dominant concept should behave coherently. &
		One concept activates on unrelated spans, making naming unreliable. \\
		
		$\mathcal{L}_{\mathrm{use}}$ &
		Class- or label-level examples should avoid repeated usage collapse. &
		Many examples rely on the same few concepts despite heterogeneous evidence. \\
		\bottomrule
	\end{tabularx}
\end{table}

For example $i$, define the flattened layer-wise concept vector
\begin{equation}
	\bar{c}_i
	=
	[
	c_{i,\mathrm{final}}^{(1)};
	\ldots;
	c_{i,\mathrm{final}}^{(L)}
	]
	\in\mathbb{R}^{D},
	\qquad
	D=LC .
\end{equation}

\subsection{Activation sparsity}

The activation sparsity loss is
\begin{equation}
	\mathcal{L}_{1}
	=
	\frac{1}{BD}
	\sum_{i=1}^{B}
	\sum_{j=1}^{D}
	|\bar{c}_{i,j}| .
	\label{eq:l1_loss}
\end{equation}
This term discourages unnecessarily large concept activations. It is most useful
when compact evidence is desirable, but we do not assume that all tasks should
produce equally sparse concept usage.

\subsection{Example-level concept exclusivity}

For each example, we normalize absolute concept activations into a usage
distribution:
\begin{equation}
	p_{i,j}
	=
	\frac{
		|\bar{c}_{i,j}|
	}{
		\sum_{m=1}^{D}|\bar{c}_{i,m}|+\epsilon
	}.
\end{equation}
The exclusivity loss is the entropy of this distribution:
\begin{equation}
	\mathcal{L}_{\mathrm{excl}}
	=
	\frac{1}{B}
	\sum_{i=1}^{B}
	\left(
	-
	\sum_{j=1}^{D}
	p_{i,j}\log(p_{i,j}+\epsilon)
	\right).
	\label{eq:exclusivity_loss}
\end{equation}
Minimizing this term encourages each example to concentrate its evidence in a
smaller subset of concept dimensions. Thus, $\mathcal{L}_{1}$ controls
activation magnitude, whereas $\mathcal{L}_{\mathrm{excl}}$ controls how
evidence is distributed across concepts.

\subsection{Gate sparsity}

The gate sparsity loss penalizes average gate activation over valid tokens:
\begin{equation}
	\mathcal{L}_{\mathrm{gate}}
	=
	\frac{1}{|\mathcal{G}|}
	\sum_{g\in\mathcal{G}}
	\mathrm{mean}_{(i,t,j):m_{i,t}=1}
	\left[
	g_{i,t,j}
	\right],
	\label{eq:gate_loss}
\end{equation}
where $\mathcal{G}$ is the set of gate tensors returned by the selected concept
adapters. In implementation, this term is computed over non-padding tokens. If a
gate tensor is unavailable or has no valid entries, it contributes zero. This
term regularizes the selection mechanism itself rather than relying only on
small-valued activations after the fact.

\subsection{Redundancy reduction}

To discourage duplicated concept dimensions, we compute the batch-level concept
matrix
\begin{equation}
	\bar{C}
	=
	[\bar{c}_1;\ldots;\bar{c}_B]
	\in\mathbb{R}^{B\times D}.
\end{equation}
After standardizing each concept dimension across the batch to obtain
$\hat{C}$, we compute
\begin{equation}
	R
	=
	\frac{1}{B}
	\hat{C}^{\top}\hat{C}.
\end{equation}
The orthogonality loss penalizes off-diagonal correlations:
\begin{equation}
	\mathcal{L}_{\mathrm{orth}}
	=
	\frac{1}{D(D-1)}
	\sum_{u\neq v}
	R_{u,v}^{2}.
	\label{eq:orth_loss}
\end{equation}
This term reduces linear duplication but does not require concept dimensions to
be statistically independent. In implementation, standardization is
$\epsilon$-stabilized, and near-constant dimensions are handled conservatively
to avoid numerical artifacts in small batches.

\subsection{Token-level concept coherence}

Because GCA produces token-level concepts before pooling, we encourage tokens
associated with the same dominant concept to have coherent concept-space
behavior. In this subsection, token sets are formed over valid tokens in the
current mini-batch, with the batch index suppressed for readability. For each
valid token, define its winning concept as
\begin{equation}
	r_t^{(\ell)}
	=
	\arg\max_{j}
	|c_{t,j}^{(\ell)}| .
\end{equation}
For concept $j$ at layer $\ell$, let $\mathcal{T}_{j}^{(\ell)}$ be the set of
valid tokens whose winning concept is $j$. If this set is sufficiently large,
its centroid is
\begin{equation}
	\mu_j^{(\ell)}
	=
	\frac{1}{|\mathcal{T}_{j}^{(\ell)}|}
	\sum_{t\in\mathcal{T}_{j}^{(\ell)}}
	c_t^{(\ell)} .
\end{equation}
The intra-cluster loss is
\begin{equation}
	\mathcal{L}_{\mathrm{intra}}
	=
	\frac{1}{|\mathcal{A}|}
	\sum_{(\ell,j)\in\mathcal{A}}
	\frac{1}{|\mathcal{T}_{j}^{(\ell)}|}
	\sum_{t\in\mathcal{T}_{j}^{(\ell)}}
	\left\|
	c_t^{(\ell)}
	-
	\mu_j^{(\ell)}
	\right\|_2^2 ,
	\label{eq:intra_loss}
\end{equation}
where $\mathcal{A}$ is the set of layer--concept pairs with enough assigned
tokens. The discrete $\arg\max$ assignments are used only as within-batch
pseudo-assignments; gradients are not intended to flow through the assignment
operation itself. If a layer--concept pair has too few assigned tokens, it is
skipped.

\subsection{Task-dependent usage diversity and usage repulsion}
\label{subsubsec:usage_diversity}

A concept model can collapse onto the same small group of concepts for many
labels or examples. To discourage this failure mode, we use a task-dependent
usage regularizer when validation ablations show that it improves the
accuracy--auditability trade-off.

For each example, define the normalized concept usage vector
\begin{equation}
	u_i
	=
	\frac{
		|\bar{c}_i|
	}{
		\sum_{j=1}^{D}|\bar{c}_{i,j}|+\epsilon
	}
	\in\mathbb{R}^{D}.
\end{equation}
For single-label classification, let
$\mathcal{B}_k=\{i:y_i=k\}$ be the examples of class $k$. For multi-label
classification, let $\mathcal{B}_k=\{i:y_{i,k}=1\}$ be examples positive for
label $k$. We define centroids only for classes or labels with enough positive
examples in the current batch. For such a class or label, the usage centroid is
\begin{equation}
	\mu_k
	=
	\frac{1}{|\mathcal{B}_k|}
	\sum_{i\in\mathcal{B}_k}
	u_i .
\end{equation}

In the full usage-diversity variant, the usage regularizer is
\begin{equation}
	\mathcal{L}_{\mathrm{use}}
	=
	\mathcal{L}_{\mathrm{ent}}
	+
	\beta_{\mathrm{within}}\mathcal{L}_{\mathrm{within}}
	+
	\beta_{\mathrm{between}}\mathcal{L}_{\mathrm{between}} .
	\label{eq:usage_div_loss}
\end{equation}
The entropy-floor term is
\begin{equation}
	\mathcal{L}_{\mathrm{ent}}
	=
	\frac{1}{|\mathcal{K}|}
	\sum_{k\in\mathcal{K}}
	\left[
	\max
	\left(
	0,
	\rho
	-
	\frac{H(\mu_k)}{\log D}
	\right)
	\right]^2 ,
	\label{eq:usage_entropy}
\end{equation}
where $\mathcal{K}$ is the set of classes or labels with enough positive
examples in the batch.

The within-label repulsion term is
\begin{equation}
	\mathcal{L}_{\mathrm{within}}
	=
	\frac{1}{|\mathcal{K}|}
	\sum_{k\in\mathcal{K}}
	\frac{1}{|\mathcal{P}_k|}
	\sum_{(i,j)\in\mathcal{P}_k}
	\left[
	\max(0,\cos(u_i,u_j)-m)
	\right]^2 ,
	\label{eq:usage_within}
\end{equation}
where $\mathcal{P}_k$ is the set of same-label example pairs for class or label
$k$, and $m$ is a cosine margin. The between-centroid term is
\begin{equation}
	\mathcal{L}_{\mathrm{between}}
	=
	\frac{1}{|\mathcal{Q}|}
	\sum_{(k,q)\in\mathcal{Q}}
	\left[
	\max(0,\cos(\mu_k,\mu_q)-m)
	\right]^2 ,
	\label{eq:usage_between}
\end{equation}
where $\mathcal{Q}$ contains pairs of distinct classes or labels present in the
batch.

For toxicity detection, we use a lighter label-conditioned usage-repulsion
variant:
\begin{equation}
	\mathcal{L}_{\mathrm{use}}
	=
	\frac{1}{|\mathcal{K}'|}
	\sum_{k\in\mathcal{K}'}
	\frac{1}{|\mathcal{P}_k|}
	\sum_{(i,j)\in\mathcal{P}_k}
	\left[
	\max(0,\cos(u_i,u_j)-m)
	\right]^2 .
	\label{eq:toxicity_usage_repulsion}
\end{equation}
Here $\mathcal{K}'$ is the set of valid toxicity labels considered by the
regularizer. The pair set $\mathcal{P}_k$ is defined as in
Eq.~\eqref{eq:usage_within}, using positive examples for toxicity label $k$.
Labels with too few positive examples in the batch are skipped. This lighter
form targets repeated top-concept collapse among positive examples of the same
label without forcing every individual example to use many concepts.

For GoEmotions and Qwen verb cloze, the final selected configurations disable
this term by setting $\lambda_{\mathrm{use}}=0$. In GoEmotions, fine-grained
emotion labels often share overlapping evidence, making usage diversity less
helpful in the selected validation trade-off. In Qwen verb cloze,
full-vocabulary token prediction naturally requires a more distributed evidence
profile, so class- or label-conditioned usage diversification is not appropriate.

\section{Additional Experimental Details}
\label{app:experimental_details}

This appendix records reproducibility details that complement
Section~\ref{sec:experimental_setup}. It includes task summaries, model variants,
training configurations, preprocessing details, selected regularization weights,
concept-labeling generation settings, and runtime information.

\begin{table}[pos=t]
	\centering
	\scriptsize
	\caption{Summary of datasets, task formulations, backbones, and evaluation roles.}
	\label{tab:task_overview}
	\begin{tabularx}{\linewidth}{p{0.14\linewidth} p{0.20\linewidth} p{0.15\linewidth} p{0.19\linewidth} X}
		\toprule
		Setting & Dataset and split & Backbone & Prediction target & Main role in evaluation \\
		\midrule
		Emotion classification &
		GoEmotions; 43,410 / 5,426 / 5,427 &
		RoBERTa-base &
		28-label sigmoid output &
		Main controlled benchmark for architecture variants, loss ablations, layer
		routing, intervention analysis, per-label behavior, and concept-name audit. \\
		
		Topic classification &
		AG News; 108,000 / 12,000 / 7,600 &
		RoBERTa-base &
		4-way softmax output &
		Direct comparison with external concept-based baselines on a compact
		single-label topic task. \\
		
		Toxicity detection &
		Jigsaw/Civil Comments; 1,443,895 / 180,487 / 180,487 &
		ModernBERT-base &
		7-label sigmoid output; 6 labels used for aggregate metrics &
		Safety-relevant multi-label setting for intervention analysis and post-hoc
		concept-level control. \\
		
		Verb cloze &
		UD English-EWT; 28,156 / 3,509 / 3,422 &
		Qwen &
		Full-vocabulary single-token prediction &
		Controlled decoder-side stress test of concept-mediated full-vocabulary
		prediction. \\
		\bottomrule
	\end{tabularx}
	
	\vspace{1mm}
	\scriptsize
	\textit{Note.} Split sizes are reported as train / validation / test. For
	toxicity detection, \textit{severe\_toxicity} is included during training but
	excluded from aggregate validation and test metrics because of insufficient
	positive validation examples under the threshold-selection split.
\end{table}

\begin{table}[pos=t]
	\centering
	\scriptsize
	\caption{Model variants, trainable components, and evaluation purpose.}
	\label{tab:model_variants}
	\begin{tabularx}{\linewidth}{p{0.16\linewidth} p{0.25\linewidth} p{0.20\linewidth} X}
		\toprule
		Variant & Backbone and trainable components & Tasks used & Purpose \\
		\midrule
		\textsc{RoBERTa}, \textsc{FT-ModernBERT}, Qwen-FT-NoConcept &
		Backbone and prediction head fine-tuned without concept adapters &
		Matched task for each backbone &
		Unconstrained predictive baselines used to measure the cost of adding a
		concept-mediated prediction path. \\
		
		\textsc{MLC} &
		Backbone parameters allowed by the task-specific fine-tuning policy,
		layer-wise GCA modules, and prediction head trained jointly &
		GoEmotions, AG News, toxicity, Qwen &
		Main concept-bearing model; tests joint learning of a layer-wise concept
		interface. \\
		
		\textsc{TI-F-MLC} &
		Task-adapted backbone frozen; GCA modules and prediction head trained &
		GoEmotions, AG News &
		Performance-preserving diagnostic variant; tests whether a task-specialized
		encoder can be wrapped with an auditable concept readout. \\
		
		\textsc{F-MLC} &
		Pretrained backbone frozen; GCA modules and prediction head trained &
		GoEmotions &
		Tests concept learning when the backbone itself is not task-adapted. \\
		
		\textsc{MLA} &
		Multi-layer adapter architecture trained with task loss only; no
		concept-structure regularizers &
		GoEmotions, AG News &
		Separates architecture and capacity from concept-oriented regularization. \\
		
		\textsc{SLC} &
		Single final-layer concept bottleneck with concept-structure regularization &
		GoEmotions &
		Tests whether a final-layer-only concept interface can replace layer-wise
		concept routing. \\
		
		\textsc{SLA} &
		Single final-layer adapter-only control trained with task loss only &
		GoEmotions &
		Controls for single-layer bottleneck capacity without concept-oriented
		regularization. \\
		
		CB-LLM, HI-Concept &
		External concept-based baselines reported or reproduced on AG News &
		AG News &
		Places GCA relative to recent concept-construction regimes on the directly
		overlapping benchmark. \\
		\bottomrule
	\end{tabularx}
\end{table}

\begin{table}[pos=t]
	\centering
	\scriptsize
	\caption{Main training configurations for the reported GCA models.}
	\label{tab:training_config_summary}
	\begin{tabularx}{\linewidth}{p{0.24\linewidth} c p{0.14\linewidth} c c c p{0.17\linewidth} c}
		\toprule
		Model & $C$ & Selected layers & $h_{\mathrm{attn}}$ & Len. & Batch & Learning rate(s) & Seeds \\
		\midrule
		GoEmotions \textsc{MLC} &
		56 & 12 encoder layers & 112 & 64 & 64 & $2{\times}10^{-5}$ & 5 \\
		
		AG News \textsc{MLC} &
		56 & 12 encoder layers & 112 & 180 & 32 & $2{\times}10^{-5}$ & 5 \\
		
		Toxicity \textsc{ModernBERT-MLC} &
		40 & all encoder layers & 80 & 256 & 64 & $2{\times}10^{-5}$ & 5 \\
		
		Qwen verb cloze &
		64 & last 12 decoder layers & 128 & 96 & 16 &
		$10^{-5}$ / $2{\times}10^{-4}$ & 3 \\
		\bottomrule
	\end{tabularx}
	
	\vspace{1mm}
	\scriptsize
	\textit{Note.} $C$ is the number of concept dimensions per selected layer.
	$h_{\mathrm{attn}}$ is the hidden size of the label-wise layer-attention head.
	For Qwen, the two learning rates correspond to unfrozen decoder parameters and
	the concept/vocabulary head, respectively. For Qwen, concepts are read from the
	last 12 decoder layers, while only the last eight decoder layers, the
	language-model head, and the GCA components are trainable. Maximum sequence
	lengths were selected from tokenized-length distributions while accounting for
	GPU memory constraints.
\end{table}

\subsection{Task, model, and training summaries}
\label{app:task_model_training_summaries}

Table~\ref{tab:task_overview} summarizes the task formulations, split sizes,
backbones, prediction targets, and evaluation roles. Table~\ref{tab:model_variants}
summarizes the evaluated model variants, and
Table~\ref{tab:training_config_summary} reports the main training
configurations used for the selected GCA models.

\subsection{Preprocessing and split details}
\label{app:preprocessing_details}

For each task, the maximum sequence length was selected after inspecting the
tokenized-length distribution on the training split while considering padding
cost and GPU memory.

\paragraph{GoEmotions}
We use the official train/dev/test split with 43,410 training examples, 5,426
validation examples, and 5,427 test examples over 28 labels. The maximum
sequence length is 64 and the batch size is 64. Thresholds are tuned per label
on the validation set from $\{0.05,0.10,\ldots,0.50\}$ using macro-F1.
Checkpoints are selected by validation macro-F1 after threshold tuning.

\paragraph{AG News}
We use the clean-v2 preprocessing pipeline with tokenizer text-pair encoding.
The resulting split sizes are 108,000 / 12,000 / 7,600 for train/validation/test.
The maximum sequence length is 180 and the batch size is 32. Separation is
handled by the tokenizer; we do not insert a literal \texttt{[SEP]} marker into
the raw text.
\begin{table}[pos=t]
	\centering
	\scriptsize
	\caption{Selected concept-regularization weights for the reported models. $C$
		is the number of concept dimensions per selected layer. A dash indicates that
		the term is not used.}
	\label{tab:selected_loss_weights_all_tasks}
	\begin{tabular}{lccccccc}
		\toprule
		Task / model & $C$ & $\lambda_{1}$ & $\lambda_{\mathrm{excl}}$
		& $\lambda_{\mathrm{orth}}$ & $\lambda_{\mathrm{intra}}$
		& $\lambda_{\mathrm{gate}}$ & $\lambda_{\mathrm{use}}$ \\
		\midrule
		GoEmotions selected \textsc{MLC}
		& 56 & 0.012 & 0.002 & 0.003 & $2{\times}10^{-5}$ & 0.0015 & 0 \\
		
		GoEmotions older sparse no-usage setting
		& 56 & 0.020 & 0.004 & 0.005 & $5{\times}10^{-5}$ & 0.0025 & 0 \\
		
		AG News \textsc{MLC}
		& 56 & 0.020 & 0.004 & 0.005 & $5{\times}10^{-5}$ & 0.0025 & 0.075 \\
		
		Toxicity \textsc{ModernBERT-MLC}
		& 40 & 0.030 & 0.0045 & 0.008 & $8{\times}10^{-5}$ & 0.0035 & 0.075 \\
		
		Qwen verb cloze
		& 64 & 0.010 & 0.003 & 0.002 & -- & 0.001 & 0 \\
		\bottomrule
	\end{tabular}
	
	\vspace{1mm}
	\scriptsize
	\textit{Note.} The older sparse no-usage GoEmotions setting is included only
	for comparison with earlier ablation runs; it is not the selected final
	\textsc{MLC} configuration. A zero weight means that the corresponding term was
	considered but is inactive in the final reported configuration.
\end{table}

\begin{table}[pos=t]
	\centering
	\scriptsize
	\caption{Additional task-specific training and model-selection settings.}
	\label{tab:selected_training_settings_all_tasks}
	\begin{tabular}{lcccc}
		\toprule
		Task / model & Backbone & Epochs & Batch & Usage margin \\
		\midrule
		GoEmotions \textsc{MLC}
		& RoBERTa-base & 15 & 64 & -- \\
		
		AG News \textsc{MLC}
		& RoBERTa-base & 5 & 32 & 0.35 \\
		
		Toxicity \textsc{ModernBERT-MLC}
		& ModernBERT-base & 2 & 64 / 32 & 0.25 \\
		
		Qwen verb cloze
		& \texttt{Qwen/Qwen2-7B-Instruct} & 2 & 16 / 4 & -- \\
		\bottomrule
	\end{tabular}
	
	\vspace{1mm}
	\scriptsize
	\textit{Note.} Batch sizes are reported as train / evaluation when they differ.
	GoEmotions uses validation-tuned per-label thresholds. Toxicity excludes
	\textit{severe\_toxicity} from aggregate metrics. Qwen uses three seeds and
	disables the usage term.
\end{table}

\paragraph{Toxicity detection}
The toxicity split contains 1,443,895 / 180,487 / 180,487 comments. The model is
trained with seven labels, but aggregate metrics exclude
\textit{severe\_toxicity} because it has too few or no positive validation
examples under the split used for threshold selection. We use target-only
positive class weighting, no weighted sampler, maximum length 256, batch size 64
for training, and batch size 32 for evaluation.

\paragraph{Qwen verb cloze}
The Qwen verb-cloze task uses \texttt{Qwen/Qwen2-7B-Instruct} as the decoder
backbone. We mask one \textsc{verb} or \textsc{aux} token in UD English-EWT,
discard examples whose target is not a single tokenizer token, and discard
prompts exceeding the maximum length. After filtering, the train/validation/test
sizes are 28,156 / 3,509 / 3,422. The maximum sequence length is 96. Full test
evaluation uses batch size 4 for memory safety.

\subsection{Selected regularization and model-selection settings}
\label{app:loss_weight_search}

Table~\ref{tab:selected_loss_weights_all_tasks} reports the selected
concept-regularization weights. The weights are task-specific because the
desired concept profile differs across fine-grained emotion classification,
topic classification, toxicity detection, and full-vocabulary token prediction.
GoEmotions receives the most detailed ablation study because it is the main
controlled benchmark. AG News uses class-conditioned usage diversity, toxicity
uses a lighter label-conditioned usage-repulsion term, and GoEmotions and Qwen
disable usage diversity in the final selected configurations.

\subsection{Concept-labeling generation settings}
\label{app:concept_labeling_generation_settings}

Concept names are assigned after training and are not used as supervision.
Table~\ref{tab:concept_labeling_generation_settings} lists the generation
settings for the candidate labelers and referee.

\begin{table}[pos=t]
	\centering
	\scriptsize
	\caption{Generation settings for post-hoc concept labeling.}
	\label{tab:concept_labeling_generation_settings}
	\begin{tabular}{lcc}
		\toprule
		Role / model & Decoding & Max. new tokens \\
		\midrule
		Mistral candidate labeler & temperature 0 & 220 \\
		Llama candidate labeler & greedy decoding & 220 \\
		Qwen referee & 3 deterministic referee votes & 140 \\
		\bottomrule
	\end{tabular}
\end{table}

We use \texttt{Mistral-7B-Instruct-v0.3} and
\texttt{meta-llama/Llama-3.1-8B-Instruct} as independent candidate labelers and
\texttt{Qwen/Qwen2-7B-Instruct} as the final referee. The referee receives the
candidate names together with the same mined evidence packet and selects the
final name used in qualitative reporting. The three deterministic referee votes
are aggregated to obtain the final referee decision.

\subsection{Runtime and hardware}
\label{app:runtime}

All experiments are run on an NVIDIA RTX 3090 GPU. The runtimes in
Table~\ref{tab:runtime_summary} are approximate and are intended for
reproducibility, not as optimized deployment benchmarks.

\begin{table}[pos=t]
	\centering
	\scriptsize
	\caption{Approximate runtime on an NVIDIA RTX 3090 GPU.}
	\label{tab:runtime_summary}
	\begin{tabular}{lcc}
		\toprule
		Task/model & Training time & Test inference time \\
		\midrule
		GoEmotions \textsc{MLC} & $\approx$2 min/epoch & $\approx$8 s \\
		AG News \textsc{MLC} & $\approx$14 min/epoch & $\approx$58 s \\
		Toxicity \textsc{ModernBERT-MLC} & $\approx$3.7 h/epoch & $\approx$10 min \\
		Qwen verb cloze & $\approx$4.7 h/epoch & $\approx$4.4 min for 3,422 examples \\
		\bottomrule
	\end{tabular}
\end{table}

For Qwen verb cloze, training follows the partial fine-tuning policy described
above and uses mixed precision when supported. Full test evaluation uses batch
size 4 and processes 856 batches in approximately 4 minutes and 26 seconds. The
reported training time is per seed; final Qwen results use three seeds for both
Qwen-\textsc{MLC} and the matched Qwen-FT-NoConcept baseline.

\section{Additional GoEmotions Analyses}
\label{app:goemotions_additional}

This appendix provides additional quantitative analyses for GoEmotions:
label-wise performance gaps, top-$k$ intervention sensitivity, local
regularization-strength sensitivity, controlled loss ablations, and a
qualitative concept-space visualization.

\subsection{Label-wise performance-gap analysis}
\label{app:goemotions_labelwise_gap}

Table~\ref{tab:goemotions_labelwise_gap_bins_app} groups labels by test-set
support and reports the mean F1 gap relative to the unconstrained
\textsc{RoBERTa} baseline. The end-to-end \textsc{MLC} pays its largest cost on
rare labels, while \textsc{TI-F-MLC} substantially reduces the gap outside the
rarest bin.

\begin{table}[pos=t]
	\centering
	\scriptsize
	\caption{Label-wise GoEmotions F1 gap by label frequency. Gaps are computed as
		model F1 minus \textsc{RoBERTa} F1 and averaged within each support bin.}
	\label{tab:goemotions_labelwise_gap_bins_app}
	\begin{tabular}{lrrrr}
		\toprule
		Support bin & Labels & Mean support & \textsc{MLC} gap & \textsc{TI-F-MLC} gap \\
		\midrule
		Rare ($<100$) & 8 & 38.8 & $-0.124$ & $-0.049$ \\
		Low (100--249) & 12 & 157.5 & $-0.046$ & $-0.0005$ \\
		Medium (250--499) & 6 & 306.3 & $-0.033$ & $+0.0017$ \\
		Frequent ($\geq 500$) & 2 & 1145.5 & $-0.029$ & $-0.0016$ \\
		\bottomrule
	\end{tabular}
	
	\vspace{1mm}
	\scriptsize
	\textit{Note.} The support--gap Spearman correlation is $0.585$ for
	\textsc{MLC} and $0.552$ for \textsc{TI-F-MLC}.
\end{table}

\begin{table}[pos=t]
	\centering
	\scriptsize
	\caption{Representative label-wise GoEmotions F1 scores for low-support labels.}
	\label{tab:goemotions_labelwise_gap_examples}
	\begin{tabular}{lrrrr}
		\toprule
		Label & Support & \textsc{RoBERTa} & \textsc{MLC} & \textsc{TI-F-MLC} \\
		\midrule
		grief & 6 & 0.239 & 0.099 & 0.077 \\
		relief & 11 & 0.259 & 0.068 & 0.187 \\
		pride & 16 & 0.496 & 0.333 & 0.458 \\
		nervousness & 23 & 0.383 & 0.263 & 0.311 \\
		embarrassment & 37 & 0.492 & 0.290 & 0.507 \\
		remorse & 56 & 0.696 & 0.674 & 0.686 \\
		fear & 78 & 0.670 & 0.642 & 0.654 \\
		desire & 83 & 0.529 & 0.405 & 0.490 \\
		excitement & 103 & 0.429 & 0.353 & 0.424 \\
		disgust & 123 & 0.483 & 0.492 & 0.473 \\
		caring & 135 & 0.439 & 0.326 & 0.419 \\
		surprise & 141 & 0.564 & 0.493 & 0.565 \\
		realization & 145 & 0.217 & 0.201 & 0.224 \\
		disappointment & 151 & 0.320 & 0.245 & 0.333 \\
		confusion & 153 & 0.449 & 0.354 & 0.445 \\
		\bottomrule
	\end{tabular}
\end{table}

Labels with extremely small support, such as \textit{grief} and \textit{relief},
are included for transparency but should not be interpreted as stable per-label
estimates. Their role is to show where the macro-F1 gap is most visible, not to
support strong conclusions about individual rare labels.

\subsection{Top-$k$ intervention sensitivity}
\label{app:goemotions_topk_ablation}

Table~\ref{tab:goemotions_mlc_topk_ablation_sensitivity} reports how the
intervention effect changes as more top-ranked label-conditioned dimensions are
removed. The targeted drop grows as the intervention removes more high-support
dimensions, while random removal remains close to zero.

\begin{table}[pos=t]
	\centering
	\scriptsize
	\caption{Top-$k$ targeted-vs-random ablation sensitivity for the selected
		GoEmotions \textsc{MLC}. Drops are selected-label probability decreases.}
	\label{tab:goemotions_mlc_topk_ablation_sensitivity}
	\begin{tabular}{lccc}
		\toprule
		Ablated dimensions & Targeted Drop & Random Drop & Gap \\
		\midrule
		Top-5 & 0.2005 $\pm$ 0.0101 & 0.0004 $\pm$ 0.0003 & 0.2001 $\pm$ 0.0100 \\
		Top-10 & 0.2827 $\pm$ 0.0167 & 0.0010 $\pm$ 0.0006 & 0.2817 $\pm$ 0.0166 \\
		Top-20 & 0.3534 $\pm$ 0.0206 & 0.0017 $\pm$ 0.0015 & 0.3517 $\pm$ 0.0200 \\
		\bottomrule
	\end{tabular}
\end{table}

This pattern shows that the support ranking is not arbitrary: the dimensions
identified as important by the model are also the dimensions whose removal
changes the prediction.

\subsection{Regularization-strength sensitivity}
\label{app:goemotions_regularization_sensitivity}

Tables~\ref{tab:goemotions_regularization_sensitivity_predictive}
and~\ref{tab:goemotions_regularization_sensitivity_concepts} report a local
3-seed sensitivity check around the selected \textsc{MLC} regularization
weights. This is not an exhaustive hyperparameter search. It is a sanity check:
moderate changes in regularization strength should not reverse the main
accuracy--auditability pattern.

\begin{table}[pos=t]
	\centering
	\scriptsize
	\caption{GoEmotions regularization-strength sensitivity: predictive metrics.}
	\label{tab:goemotions_regularization_sensitivity_predictive}
	\begin{tabular}{lcc}
		\toprule
		Setting & Macro-F1 & Micro-F1 \\
		\midrule
		Task only & 0.4511 $\pm$ 0.0113 & 0.5710 $\pm$ 0.0072 \\
		$0.5{\times}$ selected & 0.4518 $\pm$ 0.0062 & 0.5703 $\pm$ 0.0029 \\
		Selected $1.0{\times}$ & 0.4508 $\pm$ 0.0040 & 0.5728 $\pm$ 0.0089 \\
		$1.5{\times}$ selected & 0.4539 $\pm$ 0.0058 & 0.5649 $\pm$ 0.0042 \\
		Old stronger sparse & 0.4468 $\pm$ 0.0128 & 0.5634 $\pm$ 0.0060 \\
		\bottomrule
	\end{tabular}
\end{table}

\begin{table}[pos=t]
	\centering
	\scriptsize
	\caption{GoEmotions regularization-strength sensitivity: concept diagnostics
		and intervention. Higher Top-10 Mass and Gini indicate more concentrated
		concept usage; lower redundancy is better. The gap is the targeted-minus-random
		selected-label probability drop.}
	\label{tab:goemotions_regularization_sensitivity_concepts}
	\begin{tabular}{lcccc}
		\toprule
		Setting & Top-10 Mass & Gini & Redundancy & Gap \\
		\midrule
		Task only & 0.0689 $\pm$ 0.0027 & 0.5493 $\pm$ 0.0053 & 0.2483 $\pm$ 0.0055 & 0.3297 $\pm$ 0.0184 \\
		$0.5{\times}$ selected & 0.0995 $\pm$ 0.0056 & 0.6652 $\pm$ 0.0123 & 0.2285 $\pm$ 0.0017 & 0.2909 $\pm$ 0.0279 \\
		Selected $1.0{\times}$ & 0.1296 $\pm$ 0.0045 & 0.7376 $\pm$ 0.0075 & 0.2090 $\pm$ 0.0012 & 0.2911 $\pm$ 0.0295 \\
		$1.5{\times}$ selected & 0.1664 $\pm$ 0.0061 & 0.7949 $\pm$ 0.0083 & 0.1893 $\pm$ 0.0026 & 0.2954 $\pm$ 0.0287 \\
		Old stronger sparse & 0.1738 $\pm$ 0.0037 & 0.8095 $\pm$ 0.0051 & 0.1864 $\pm$ 0.0070 & 0.2923 $\pm$ 0.0411 \\
		\bottomrule
	\end{tabular}
	
	\vspace{1mm}
	\scriptsize
	\textit{Note.} These 3-seed sensitivity runs are not the official 5-seed test
	estimate for the selected model. Differences from the main tables reflect seed
	composition and rerun variance. The selected weights are
	$\lambda_1=0.012$, $\lambda_{\mathrm{excl}}=0.002$,
	$\lambda_{\mathrm{orth}}=0.003$,
	$\lambda_{\mathrm{intra}}=2{\times}10^{-5}$, and
	$\lambda_{\mathrm{gate}}=0.0015$, with usage diversity disabled. The older
	sparse setting uses
	$0.020/0.004/0.005/5{\times}10^{-5}/0.0025$ for the same terms.
\end{table}

\subsection{Loss ablations for the multi-layer concept model}
\label{app:goemotions_loss_ablation}

Tables~\ref{tab:goemotions_loss_ablation_predictive} and
\ref{tab:goemotions_loss_ablation_concepts} report a controlled 5-seed
loss-ablation rerun for the GoEmotions multi-layer concept model. All rows use
the same architecture, official split, validation-tuned per-label threshold
protocol, and model-selection procedure. The selected configuration is chosen as
a balanced accuracy--auditability trade-off, not as the maximizer of every
metric.

\begin{table}[pos=t]
	\centering
	\scriptsize
	\caption{GoEmotions loss ablations: predictive metrics. Values are 5-seed test
		means $\pm$ standard deviations from a controlled ablation rerun.}
	\label{tab:goemotions_loss_ablation_predictive}
	\begin{tabular}{lccc}
		\toprule
		Configuration & Macro-F1 & Micro-F1 & Macro-AUPRC \\
		\midrule
		Task-only bottleneck & 0.4503 $\pm$ 0.0154 & 0.5623 $\pm$ 0.0074 & 0.4292 $\pm$ 0.0079 \\
		Selected + usage-div & 0.4422 $\pm$ 0.0126 & 0.5578 $\pm$ 0.0096 & 0.4100 $\pm$ 0.0112 \\
		\textbf{Selected \textsc{MLC}} & 0.4531 $\pm$ 0.0124 & 0.5682 $\pm$ 0.0060 & 0.4291 $\pm$ 0.0058 \\
		No $L_1$ & 0.4546 $\pm$ 0.0093 & 0.5678 $\pm$ 0.0016 & 0.4303 $\pm$ 0.0088 \\
		No exclusivity & 0.4618 $\pm$ 0.0124 & 0.5696 $\pm$ 0.0070 & 0.4336 $\pm$ 0.0073 \\
		No orthogonality & 0.4550 $\pm$ 0.0081 & 0.5674 $\pm$ 0.0067 & 0.4315 $\pm$ 0.0075 \\
		No intra-cluster & 0.4583 $\pm$ 0.0109 & 0.5670 $\pm$ 0.0061 & 0.4338 $\pm$ 0.0091 \\
		No gate sparsity & 0.4551 $\pm$ 0.0079 & 0.5665 $\pm$ 0.0030 & 0.4313 $\pm$ 0.0082 \\
		\bottomrule
	\end{tabular}
	
	\vspace{1mm}
	\scriptsize
	\textit{Note.} This table is a controlled ablation rerun. Small differences
	from the main selected-model table reflect independent reruns under the same
	model-selection and threshold-selection protocol.
\end{table}

\begin{table}[pos=t]
	\centering
	\scriptsize
	\caption{GoEmotions loss ablations: concept diagnostics and intervention.
		Higher Top-10 Mass and Gini indicate more concentrated concept usage; lower
		redundancy indicates less correlated concept dimensions. The ablation gap is the
		targeted-minus-random selected-label probability drop under top-10 concept
		ablation.}
	\label{tab:goemotions_loss_ablation_concepts}
	\begin{tabular}{lcccc}
		\toprule
		Configuration & Top-10 Mass & Gini & Redundancy & Ablation Gap \\
		\midrule
		Task-only bottleneck & 0.0681 $\pm$ 0.0026 & 0.5495 $\pm$ 0.0048 & 0.2500 $\pm$ 0.0067 & 0.3230 $\pm$ 0.0106 \\
		Selected + usage-div & 0.1622 $\pm$ 0.0084 & 0.7659 $\pm$ 0.0115 & 0.2020 $\pm$ 0.0039 & 0.2946 $\pm$ 0.0206 \\
		\textbf{Selected \textsc{MLC}} & 0.1300 $\pm$ 0.0026 & 0.7413 $\pm$ 0.0077 & 0.2079 $\pm$ 0.0023 & 0.2866 $\pm$ 0.0200 \\
		No $L_1$ & 0.0758 $\pm$ 0.0027 & 0.6006 $\pm$ 0.0051 & 0.2279 $\pm$ 0.0047 & 0.3236 $\pm$ 0.0137 \\
		No exclusivity & 0.1195 $\pm$ 0.0028 & 0.7078 $\pm$ 0.0047 & 0.2142 $\pm$ 0.0043 & 0.2769 $\pm$ 0.0219 \\
		No orthogonality & 0.1297 $\pm$ 0.0033 & 0.7381 $\pm$ 0.0074 & 0.2201 $\pm$ 0.0028 & 0.2858 $\pm$ 0.0186 \\
		No intra-cluster & 0.1296 $\pm$ 0.0014 & 0.7417 $\pm$ 0.0048 & 0.2077 $\pm$ 0.0058 & 0.2852 $\pm$ 0.0219 \\
		No gate sparsity & 0.1271 $\pm$ 0.0027 & 0.7343 $\pm$ 0.0072 & 0.2097 $\pm$ 0.0021 & 0.2808 $\pm$ 0.0168 \\
		\bottomrule
	\end{tabular}
	
	\vspace{1mm}
	\scriptsize
	\textit{Note.} Random ablation drops are small across all rows
	($0.0010$--$0.0023$ on average), so the reported gap mainly reflects targeted
	concept removal.
\end{table}

\subsection{Concept-space visualization}
\label{app:goemotions_visualization}

\begin{figure}[pos=t]
	\centering
	\includegraphics[width=0.48\linewidth]{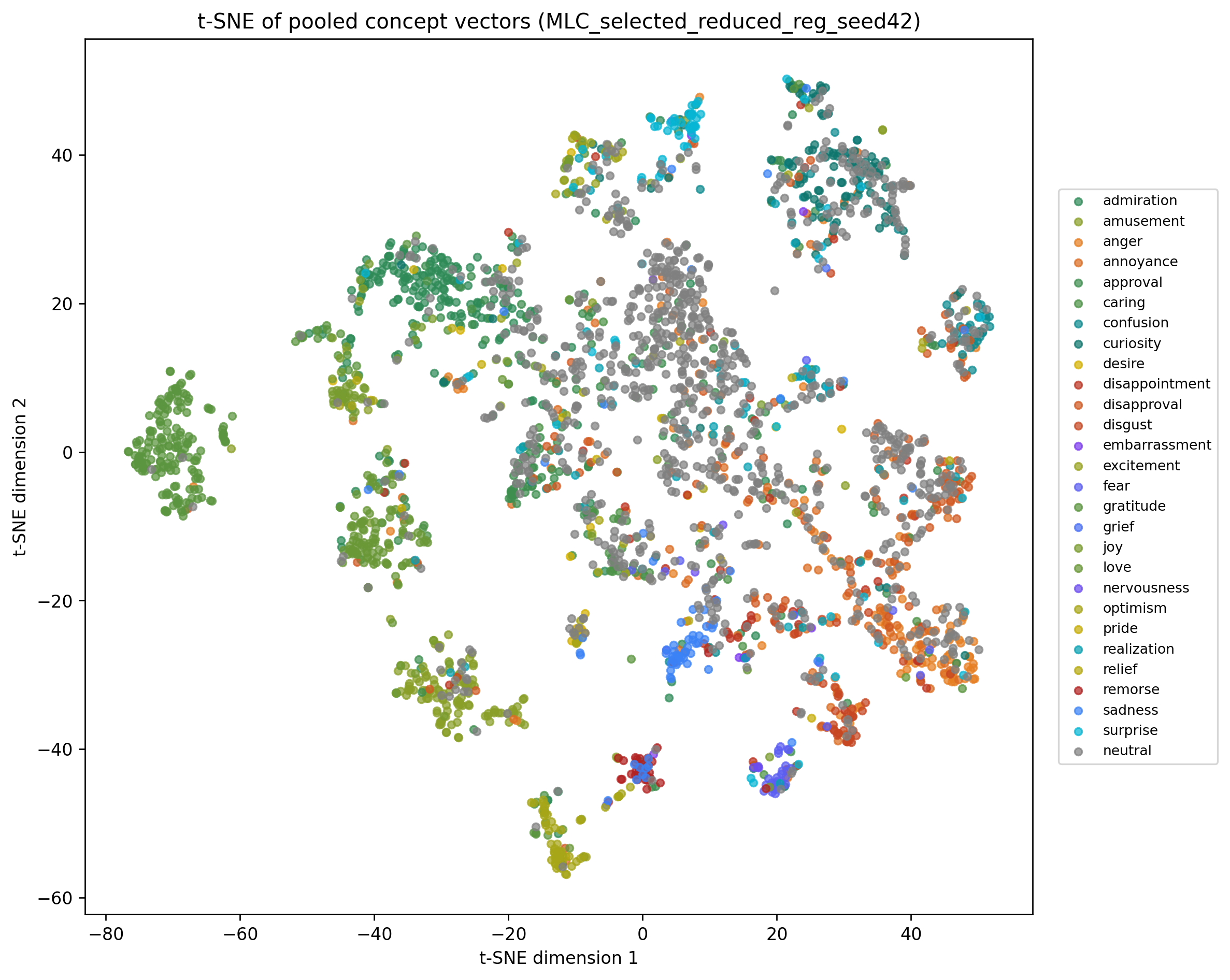}
	\includegraphics[width=0.48\linewidth]{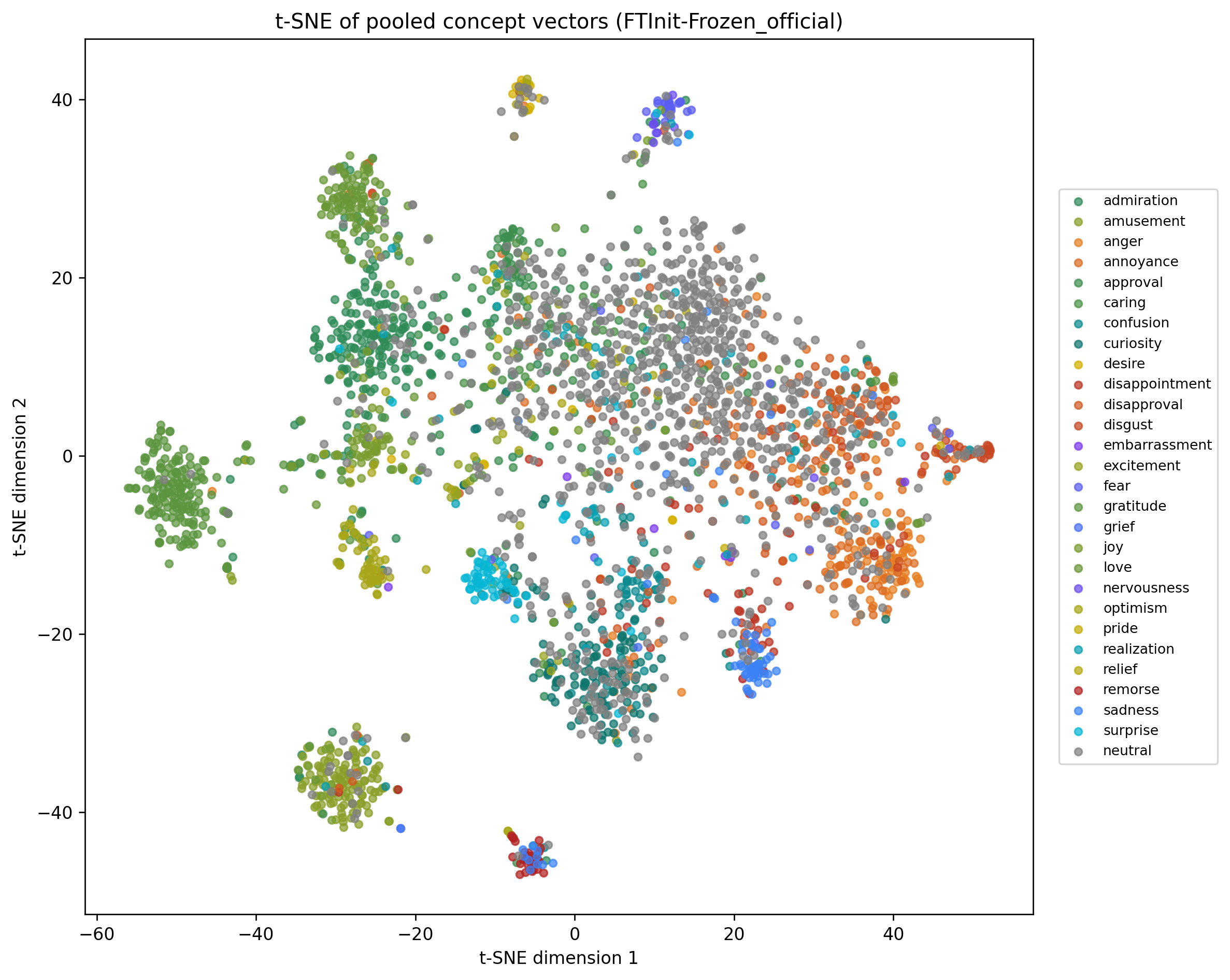}
	\caption{t-SNE projection of pooled concept vectors on GoEmotions. Left:
		selected \textsc{MLC}. Right: task-initialized frozen multi-layer concept
		model. The plot is used as qualitative visualization only.}
	\label{fig:goemotions_tsne_concepts}
\end{figure}

Figure~\ref{fig:goemotions_tsne_concepts} shows t-SNE projections of pooled
concept vectors for \textsc{MLC} and \textsc{TI-F-MLC}. The projection is
included only as a qualitative visualization and should not be interpreted as a
quantitative clustering metric.

\section{Additional Qualitative Results}
\label{app:additional_qualitative_results}

This appendix provides qualitative examples that complement the aggregate
diagnostics in Section~\ref{sec:results}. The examples are illustrative and are
not used as standalone evidence of interpretability. Concept names are post-hoc
descriptors of mined textual evidence, not ground-truth semantic annotations.

\subsection{GoEmotions concept evidence and local interventions}
\label{app:goemotions_qualitative}

Table~\ref{tab:goemotions_label_conditioned_top_concepts} reports representative
label-conditioned concept evidence for \textsc{MLC}. The ranking is computed
from the trained classifier, while the displayed names come from the
task-agnostic Qwen-refereed naming pipeline. Thus, the table separates
label-conditioned importance from post-hoc naming.

\begin{table}[pos=t]
	\centering
	\scriptsize
	\caption{Representative label-conditioned concept evidence for GoEmotions in
		\textsc{MLC}. Names are post-hoc task-agnostic summaries of mined textual
		evidence.}
	\label{tab:goemotions_label_conditioned_top_concepts}
	\begin{tabularx}{\linewidth}{p{0.12\linewidth} p{0.35\linewidth} X}
		\toprule
		Label & Representative high-ranking concepts & Mined textual evidence \\
		\midrule
		\textit{amusement}
		& \textit{Humorous}; \textit{Insulting Humor}; \textit{Making Something Humorous}; \textit{Positive Emotions}
		& ``funny'', ``fun'', ``hilarious'', ``a joke'', ``hahaha'', ``making'', ``have fun'' \\
		
		\textit{love}
		& \textit{Love For A Name}; \textit{Affection For A Name}; \textit{Express Affection Towards A Person}; \textit{Emotional Expressions}
		& ``i love this'', ``i love name'', ``i like name'', ``my favorite'', ``loves'', ``cute'', ``amazing'' \\
		
		\textit{fear}
		& \textit{Negative Emotion}; \textit{Fear}; \textit{Fear Vs Positive Emotions}; \textit{Danger}
		& ``terrible'', ``afraid'', ``scared'', ``scary'', ``horrible'', ``fear'', ``nightmare'', ``dangerous'' \\
		
		\textit{remorse}
		& \textit{Apology}; \textit{Apology And Negative Emotion}; \textit{Wishing}
		& ``i'm sorry'', ``i'm so sorry'', ``wish'', ``regret'', ``shame'', ``absolute worst'' \\
		
		\textit{optimism}
		& \textit{Aspirational Vs Negative Outlook}; \textit{Excitement Vs Hope Negation}; \textit{Hope And Luck Vs Uncertainty}; \textit{Positive Emotions}
		& ``hope'', ``i hope'', ``hopefully'', ``luck'', ``good luck'', ``you're ok'', ``goes well'' \\
		
		\textit{disgust}
		& \textit{Expressing Disappointment}; \textit{Apology And Negative Emotion}; \textit{Negative Evaluation}; \textit{Desire Vs Hate}
		& ``disgusting'', ``i hate'', ``weird'', ``nasty'', ``shame'', ``absolute'', ``the worst'', ``awful'' \\
		\bottomrule
	\end{tabularx}
\end{table}

The table shows recognizable regularities, including humor markers for
\textit{amusement}, affection/name-related expressions for \textit{love},
fear-related terms for \textit{fear}, apology cues for \textit{remorse}, and
negative-evaluation patterns for \textit{disgust}. It should still be read
cautiously: a concept name summarizes recurring evidence for a latent dimension,
whereas the label-conditioned score measures how strongly that dimension
supports a particular prediction.

\begin{table}[pos=t]
	\centering
	\scriptsize
	\caption{Representative example-level concept interventions for GoEmotions in
		\textsc{MLC}. For each explained label, the top supporting concepts are ablated
		and the label probability is recomputed.}
	\label{tab:goemotions_example_interventions}
	\begin{tabularx}{\linewidth}{p{0.04\linewidth} p{0.30\linewidth} p{0.25\linewidth} X}
		\toprule
		Case & Input snippet & Explained label and intervention & Main supporting concept evidence \\
		\midrule
		1
		& ``Lol! But I love your last name though. XD''
		& \textit{amusement}: $0.869 \rightarrow 0.113$, $\Delta=0.757$;
		\textit{love}: $0.867 \rightarrow 0.129$, $\Delta=0.739$
		& Humor-related concepts support \textit{amusement}; name/affection concepts
		support \textit{love}. \\
		
		2
		& ``I've also heard that intriguing but also kinda scary''
		& \textit{fear}: $0.870 \rightarrow 0.579$, $\Delta=0.291$
		& Fear and danger concepts dominate. \\
		
		3
		& ``I'm sorry to hear that friend :(. It's for the best most likely if she
		didn't accept you for who you are''
		& \textit{remorse}: $0.657 \rightarrow 0.366$, $\Delta=0.292$
		& Apology-related concepts dominate. \\
		
		4
		& ``Kings fan here, good luck to you guys! Will be an interesting game to
		watch!''
		& predicted \textit{optimism}: $0.724 \rightarrow 0.530$, $\Delta=0.194$
		& Hope/luck concepts support an adjacent positive-affect prediction. \\
		
		5
		& ``The thought of shooting anything at asylum seekers is appalling.''
		& predicted \textit{disgust}: $0.443 \rightarrow 0.056$, $\Delta=0.388$;
		gold \textit{fear}: $0.096 \rightarrow 0.032$, $\Delta=0.064$
		& The model routes ``appalling'' mainly through broad negative-evaluation
		concepts, supporting \textit{disgust} more strongly than the gold
		\textit{fear} label. \\
		\bottomrule
	\end{tabularx}
\end{table}

\subsection{AG News qualitative examples}
\label{app:agnews_qualitative}

Table~\ref{tab:app_agnews_interventions} reports representative AG News
example-level interventions. The examples are consistent with the aggregate
diagnostics in Table~\ref{tab:agnews_interpretability}: high-support concept
packets recover recognizable class evidence such as geopolitical conflict for
World, teams and scores for Sports, profit and market cues for Business, and
technology entities for Sci/Tech.

\begin{table}[pos=t]
	\centering
	\scriptsize
	\caption{Representative AG News example-level concept interventions for
		\textsc{MLC}. Concept names are post-hoc descriptors of mined evidence.}
	\label{tab:app_agnews_interventions}
	\begin{tabularx}{\linewidth}{p{0.10\linewidth} X p{0.30\linewidth} p{0.13\linewidth}}
		\toprule
		Pred. class & Input snippet & Ablated concept descriptors & Probability change \\
		\midrule
		Sports &
		Giddy Phelps Touches Gold for First Time. Michael Phelps won the gold medal
		in the 400 individual medley and set a world record. &
		Sports; Olympic/athlete context; teams; event-specific evidence &
		$0.946 \rightarrow 0.161$ $(\Delta=0.786)$ \\
		
		World &
		Iraqi oil exports slump: report. Near daily attacks on pipelines and pumping
		stations pushed down Iraq's oil exports to their lowest point in nearly a year. &
		Conflict; military action; Middle Eastern conflict; geopolitical region &
		$0.854 \rightarrow 0.333$ $(\Delta=0.521)$ \\
		
		Business &
		Stanley Leisure announced a plan to develop a casino complex on land adjacent
		to Leeds United's Elland Road. &
		Economic activity; financial gain; price/profit-related business evidence &
		$0.735 \rightarrow 0.384$ $(\Delta=0.350)$ \\
		
		Sci/Tech &
		Czech Republic's cell operators were fined for breaching competition rules,
		officials said Thursday. &
		Technology products; communications operators; technology source; named entities &
		$0.783 \rightarrow 0.207$ $(\Delta=0.576)$ \\
		\bottomrule
	\end{tabularx}
\end{table}

\subsection{Toxicity qualitative examples}
\label{app:toxicity_qualitative}

Table~\ref{tab:toxicity_qual_examples} reports local concept-level examples from
a representative seed-45 \textsc{ModernBERT-MLC} checkpoint whose concept labels
were generated from the same seed. For each input, layer--concept dimensions are
ranked by signed contribution to the selected label. The displayed names are
Qwen-refereed post-hoc labels and are used only for reporting.

\begin{table}[pos=t]
	\centering
	\scriptsize
	\caption{Representative qualitative examples for \textsc{ModernBERT-MLC}.}
	\label{tab:toxicity_qual_examples}
	\begin{tabularx}{\linewidth}{p{0.13\linewidth}p{0.29\linewidth}p{0.22\linewidth}X}
		\toprule
		Case & Input excerpt & Gold / prediction & Key concept evidence \\
		\midrule
		TP--direct insult
		& ``I don't deal with idiots.''
		& Gold: toxic, insult. \newline
		Pred.: toxic, $p_{\mathrm{target}}=0.9926$; insult, $p=0.8802$.
		& \emph{Insulting or derogatory comments}; \emph{Negative evaluation of a
			person}; \emph{Insulting or Offensive Language}. \\
		
		FP--rhetorical criticism
		& ``What a load of rubbish! ...''
		& Gold: non-toxic. \newline
		Pred.: toxic, $p_{\mathrm{target}}=0.9345$; insult, $p=0.4860$.
		& Concepts associated with negative or dismissive language. \\
		
		FN--identity/topic framing
		& ``Islamic Terrorism was not born ...''
		& Gold: toxic, identity\_attack. \newline
		Pred.: non-toxic, $p_{\mathrm{target}}=0.6762$ below threshold;
		identity\_attack, $p=0.1563$.
		& Identity/topic-related evidence appears, but the final probabilities remain
		below threshold. \\
		\bottomrule
	\end{tabularx}
\end{table}

The examples illustrate both value and limitations. Direct derogatory language
is captured cleanly, strongly dismissive rhetoric can lead to false positives,
and identity/topic framing can remain difficult when the model does not treat
the mention as a direct attack.

\subsection{Qwen verb-cloze qualitative examples}
\label{app:qwen_qualitative}

For Qwen verb cloze, fixed prompt tokens such as instruction headers are
excluded from evidence mining. Evidence is extracted only from the sentence span
containing the masked verb, avoiding prompt artifacts such as
\texttt{Sentence:} or \texttt{Missing verb:}.

\begin{table}[pos=t]
	\centering
	\scriptsize
	\caption{Qualitative examples from Qwen-\textsc{MLC} verb-cloze prediction.
		Evidence is extracted only from the sentence span.}
	\label{tab:qwen_verb_qual}
	\begin{tabularx}{\linewidth}{p{0.13\linewidth}p{0.35\linewidth}p{0.25\linewidth}X}
		\toprule
		Case & Sentence & Top predictions & Representative concept evidence \\
		\midrule
		Correct \textsc{aux}
		& We have utilized Mr. Pozza and his firm twice now in our family and both
		times have \texttt{[BLANK]} very pleased.
		& Gold: \textit{been}. \newline
		Top-5: \textit{been}, \textit{was}, \textit{felt}, \textit{being},
		\textit{kept}.
		& \textsc{have}-perfect frame; local evidence: ``both times have
		\texttt{[BLANK]}''. \\
		
		Plausible near miss
		& Make sure to put OILY sauces on your food to \texttt{[BLANK]} it moist!
		& Gold: \textit{make}. \newline
		Top-5: \textit{keep}, \textit{make}, \textit{save}, \textit{get},
		\textit{help}.
		& Transitive \texttt{[BLANK]}--\textit{it} frame; local evidence:
		``\texttt{[BLANK]} it moist''. \\
		
		Hard lexical error
		& But because we do not want to \texttt{[BLANK]} hateful, we must pretend
		everyone is a possible suspect.
		& Gold: \textit{sound}. \newline
		Top-5: \textit{be}, \textit{feel}, \textit{have}, \textit{continue},
		\textit{get}.
		& Infinitival verb slot; local evidence: ``want to \texttt{[BLANK]} hateful''. \\
		\bottomrule
	\end{tabularx}
\end{table}

The examples illustrate how Qwen-\textsc{MLC} can capture the local grammatical
frame. In the auxiliary example, it identifies a \textsc{have}-perfect context
and ranks the correct token first. In the near-miss example, it recognizes a
transitive \texttt{[BLANK]}--object construction and proposes plausible verbs,
although the exact gold verb is not ranked first. In the harder lexical case, it
recognizes an infinitival verb slot but prefers a generic verb over the more
specific gold token. This is consistent with the quantitative results:
concept-mediated prediction preserves some syntactic and local-contextual
information, but exact lexical selection remains harder.


\section{Reliability Audit of Post-hoc Concept Naming}
\label{app:concept_labeling_reliability}

Concept names in GCA are assigned post hoc from mined textual evidence and are
used as human-readable audit aids rather than ground-truth semantic annotations.
We therefore evaluate the naming step separately from predictive performance and
intervention metrics. The goal is not to show that every latent dimension is a
perfectly clean human concept, but to test whether the assigned name is
supported by the evidence shown to the automated auditor.

For each audited concept packet, the automated auditor receives the final
proposed concept name, top activating $n$-grams, contrastive $n$-grams from the
opposite activation direction, and representative examples when available. Task
labels, gold-label associations, predicted-label associations,
classifier-weight associations, and label-conditioned concept scores are hidden.
The audit therefore evaluates name grounding in the displayed evidence, not
downstream usefulness for a particular class.

Each concept is scored on three 1--5 scales: \emph{groundedness},
\emph{coherence}, and \emph{specificity}. A concept name is treated as
\emph{supported} if groundedness $\geq 4$, coherence $\geq 4$, and specificity
$\geq 3$; \emph{partially supported} if groundedness $\geq 3$ and coherence
$\geq 3$ but the stricter criterion is not met; \emph{unsupported} if
groundedness $\leq 2$; and \emph{unclear} otherwise. Strict acceptance is
equivalent to the supported criterion. Relaxed acceptance requires groundedness
$\geq 4$ and coherence $\geq 3$. These criteria evaluate evidence-name
alignment, not whether the concept is causally sufficient for a downstream
prediction.

For each of GoEmotions, AG News, and toxicity detection, we audit the same 100
sampled concept packets with three independent automated audit passes. A concept
is accepted if at least two of the three audits accept it under the
corresponding criterion. Table~\ref{tab:concept_labeling_reliability} reports
majority-consensus acceptance rates, and score means are pooled over all 300
individual ratings per task.

\begin{table}[pos=t]
	\centering
	\scriptsize
	\caption{Reliability audit of post-hoc concept naming. Strict and relaxed
		acceptance are concept-level majority-consensus rates over three automated
		audits of the same 100 concepts per task. Score means are pooled over 300
		individual ratings per task.}
	\label{tab:concept_labeling_reliability}
	\begin{tabular}{llcrrrrr}
		\toprule
		Task & Model & $n$ & Strict & Relaxed & Ground. & Coher. & Spec. \\
		\midrule
		GoEmotions & \textsc{MLC} & 100
		& 65.0\% & 69.0\% & 3.81 & 3.83 & 3.73 \\
		AG News & \textsc{MLC} & 100
		& 68.0\% & 76.0\% & 4.19 & 4.24 & 4.03 \\
		Toxicity & \textsc{ModernBERT-MLC} & 100
		& 62.0\% & 69.0\% & 4.07 & 4.02 & 3.85 \\
		\midrule
		Overall & -- & 300
		& 65.0\% & 71.3\% & 4.02 & 4.03 & 3.87 \\
		\bottomrule
	\end{tabular}
\end{table}

Inter-audit agreement for the same automated audit passes is reported in
Table~\ref{tab:concept_labeling_agreement}.

\begin{table}[pos=t]
	\centering
	\scriptsize
	\caption{Inter-audit agreement for post-hoc concept naming. Fleiss'
		$\kappa$ is computed over the three automated audit passes for strict
		acceptance, relaxed acceptance, and the four-way derived decision
		(\emph{supported}, \emph{partially supported}, \emph{unsupported},
		\emph{unclear}).}
	\label{tab:concept_labeling_agreement}
	\begin{tabular}{lccc}
		\toprule
		Task & Strict $\kappa$ & Relaxed $\kappa$ & Four-way $\kappa$ \\
		\midrule
		GoEmotions & 0.666 & 0.593 & 0.562 \\
		AG News & 0.641 & 0.588 & 0.536 \\
		Toxicity & 0.467 & 0.511 & 0.354 \\
		\bottomrule
	\end{tabular}
\end{table}

The audit results support a cautious interpretation of the naming pipeline. A
majority of names are strictly accepted under conservative criteria, and a
larger share are accepted under the relaxed criterion. Many of the remaining
cases are not arbitrary or meaningless; they are often partially grounded but too
broad, too narrow, attached to the wrong activation pole, or based on mixed
evidence. Agreement is stronger for GoEmotions and AG News than for toxicity,
where concept packets often combine identity mentions, pragmatic context, and
abusive-language cues.

We therefore use concept names as compact evidence-grounded descriptors, not as
semantic ground truth. The audit evaluates whether the displayed evidence
supports the assigned name, while targeted-versus-random interventions evaluate
whether the underlying dimensions are functionally used by the prediction head.

Table~\ref{tab:concept_labeling_failure_examples} illustrates representative
non-strictly accepted cases.

\begin{table}[pos=t]
	\centering
	\scriptsize
	\caption{Representative non-strictly accepted post-hoc concept names. The table
		illustrates common name-grounding failure modes; it does not evaluate model
		prediction correctness.}
	\label{tab:concept_labeling_failure_examples}
	\begin{tabularx}{\linewidth}{p{0.12\linewidth} p{0.24\linewidth} p{0.15\linewidth} X}
		\toprule
		Task & Proposed name & Failure mode & Safer reading of the evidence \\
		\midrule
		GoEmotions &
		\textit{Cake Day Celebration} &
		Too narrow &
		The evidence includes ``happy cake day'' and ``cake day'', but also broader
		wish/affection terms such as ``wish'', ``heart'', and ``home''. \\
		
		GoEmotions &
		\textit{Rock Beats} &
		Wrong pole / contrastive fragment &
		The rock--paper--scissors phrase appears mainly in contrastive evidence, while
		the primary side contains ``welcome'', ``agree'', ``support'', ``congrats'', and
		thanks-like snippets. \\
		
		AG News &
		\textit{Oil} &
		Partially grounded but too narrow &
		Some $n$-grams mention oil and prices, but examples also involve welfare
		protests, power shortages, government, and an airport collapse. \\
		
		AG News &
		\textit{Politics} &
		Contradicted by dominant evidence &
		The dominant evidence includes sports and injury cues such as ``points'',
		``knee'', ``Manning'', ``Champions League'', and ``Manchester United''. \\
		
		Toxicity &
		\textit{Negative Moral Character / thank} &
		Mixed unrelated cues &
		Several examples express thanks or short positive acknowledgements, so the
		``negative moral character'' component is not supported by the primary evidence. \\
		
		Toxicity &
		\textit{Possessive Noun Phrase / Anti-Muslim sentiment} &
		Over-interprets contrastive side &
		The primary evidence is mostly crude sexual/body-part language, while
		identity-related content appears mainly in contrastive evidence. \\
		\bottomrule
	\end{tabularx}
\end{table}

These examples reinforce the central reporting convention of the paper:
post-hoc names help humans inspect the concept space, but they are not treated
as ground-truth semantic annotations.

\printcredits

\section*{Declaration of competing interest}
The authors declare that they have no known competing financial interests or
personal relationships that could have appeared to influence the work reported
in this paper.

\section*{Funding}
This research did not receive any specific grant from funding agencies in the
public, commercial, or not-for-profit sectors.

\section*{Data availability}
The datasets used in this study are publicly available. GoEmotions, AG News,
Jigsaw/Civil Comments, and UD English-EWT are cited in the paper. Code,
configuration files, and processed experiment outputs will be released in a
public repository upon publication. An anonymized repository can be provided
during review if required by the journal.

\section*{Declaration of generative AI and AI-assisted technologies in the
	manuscript preparation process}
During the preparation of this work, the authors used OpenAI's ChatGPT to
support language polishing, organization, and clarity checking. In addition,
instruction-tuned language models were used as part of the explicitly described
post-hoc concept-label proposal and referee pipeline. After using these tools,
the authors reviewed and edited the content as needed and take full
responsibility for the content of the published article.
\bibliographystyle{cas-model2-names}

\bibliography{cas-refs}



\end{document}